\documentclass[10pt,twocolumn,letterpaper]{article}

\usepackage{cvpr}
\usepackage{times}
\usepackage{epsfig}
\usepackage{graphicx}
\usepackage{color}
\usepackage{amsmath}
\usepackage{amssymb}
\usepackage{subcaption}

\usepackage[pagebackref=true,breaklinks=true,letterpaper=true,colorlinks,bookmarks=false]{hyperref}

\cvprfinalcopy %

\def\cvprPaperID{3336} %

\usepackage{bbm}
\usepackage{booktabs}
\usepackage{multirow}

\usepackage{algorithm}
\usepackage{algpseudocode}

\newcommand{\PAR}[1]{\vskip1pt \noindent {\bf #1~}}
\newcommand{\PARbegin}[1]{\noindent {\bf #1~}}

\newcounter{row}

\newenvironment{imgrows}[1][\textwidth]%
  {\begin{minipage}{#1}%
   \setcounter{row}{0}%
   \stepcounter{figure}%
  }%
  {\addtocounter{figure}{-1}%
   \end{minipage}%
  }
\newcommand\imgrow
  {\par\noindent
   \refstepcounter{row}%
   \makebox[0.5em][r]{(\alph{row})}\
  }

\ifcvprfinal\fi
\begin{document}

\title{MOTS: Multi-Object Tracking and Segmentation}

\author{
  \hspace{-1.3cm}
  \begin{tabular}[t]{c}
    Paul Voigtlaender$^1$ \quad Michael Krause$^1$ \quad Aljo\u{s}a O\u{s}ep$^1$ \quad Jonathon Luiten$^1$\\
    Berin Balachandar Gnana Sekar$^1$\quad Andreas Geiger$^2$\quad Bastian Leibe$^1$ \\
    $^1$RWTH Aachen University \quad $^2$MPI for Intelligent Systems and University of T\"ubingen\\
    {\tt\small \{voigtlaender,osep,luiten,leibe\}@vision.rwth-aachen.de}\\
    {\tt\small \{michael.krause,berin.gnana\}@rwth-aachen.de andreas.geiger@tue.mpg.de}
\end{tabular}
}

\maketitle
\thispagestyle{empty}

\begin{abstract}
This paper extends the popular task of multi-object tracking to multi-object tracking and segmentation (MOTS). Towards this goal, we create dense pixel-level annotations for two existing tracking datasets using a semi-automatic annotation procedure. Our new annotations comprise 65,213 pixel masks for 977 distinct objects (cars and pedestrians) in 10,870 video frames. For evaluation, we extend existing multi-object tracking metrics to this new task. Moreover, we propose a new baseline method which jointly addresses detection, tracking, and segmentation with a single convolutional network. We demonstrate the value of our datasets by achieving improvements in performance when training on MOTS annotations. We believe that our datasets, metrics and baseline will become a valuable resource towards developing multi-object tracking approaches that go beyond 2D bounding boxes. We make our annotations, code, and models available at \url{https://www.vision.rwth-aachen.de/page/mots}.
\end{abstract}

\vspace{-12pt}
\section{Introduction}
\vspace{-4pt}
In recent years, the computer vision community has made significant advances in increasingly difficult tasks. Deep learning techniques now demonstrate impressive performance in object detection as well as image and instance segmentation. Tracking, on the other hand, remains challenging, especially when multiple objects are involved.
In particular, results of recent tracking evaluations \cite{Milan16Arxiv,DAVIS2018,Kristan16TPAMI} show that bounding box level tracking performance is saturating. Further improvements will only be possible when moving to the pixel level.
We thus propose to think of all three tasks -- detection, segmentation and tracking -- as interconnected problems that need to be considered together.

\begin{figure}[t!]
	\centering
		\hspace{-5pt}
		\includegraphics[width=0.09\textwidth]{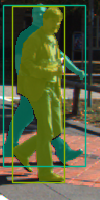}
		\includegraphics[width=0.09\textwidth]{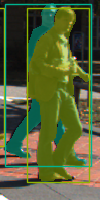}
		\vspace{1.2pt}
		\includegraphics[width=0.09\textwidth]{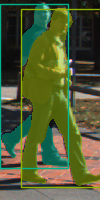}
		\includegraphics[width=0.09\textwidth]{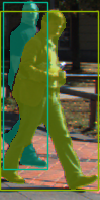}
		\\
		
		\includegraphics[width=0.09\textwidth]{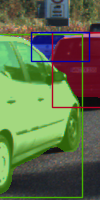}
		\includegraphics[width=0.09\textwidth]{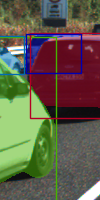}
		\includegraphics[width=0.09\textwidth]{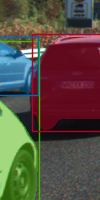}
		\includegraphics[width=0.09\textwidth]{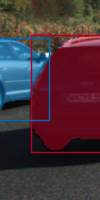}
    \vspace{-6pt}%
	\caption{\textbf{Segmentations vs. Bounding Boxes.} When objects pass each other, large parts of an object's bounding box may belong to another instance, while per-pixel segmentation masks locate objects precisely. The shown annotations are crops from our KITTI MOTS dataset.
	}
	\label{fig:people-overlap}
\end{figure}

Datasets that can be used to train and evaluate models for instance segmentation usually do not provide annotations on video data or even information on object identities across different images. Common datasets for multi-object tracking, on the other hand, provide only bounding box annotations of objects. These can be too coarse, \eg, when objects are partially occluded such that their bounding box contains more information from other objects than from themselves, see Fig.~\ref{fig:people-overlap}. In these cases, pixel-wise segmentation of the objects results in a more natural description of the scene and may provide additional information for subsequent processing steps. %
For segmentation masks there is a well-defined ground truth, whereas many different (non-tight) boxes might roughly fit an object. %
Similarly, tracks with overlapping bounding boxes create ambiguities when compared to ground truth that usually need to be resolved at evaluation time by heuristic matching procedures. Segmentation based tracking results, on the other hand, are by definition non-overlapping and can thus be compared to ground truth in a straightforward manner.

In this paper, we therefore propose to extend the well-known multi-object tracking task to instance segmentation tracking. We call this new task ``Multi-Object Tracking and Segmentation (MOTS)''. To the best of our knowledge, there exist no datasets for this task to date.
While there are many methods for bounding box tracking in the literature, MOTS requires combining temporal and mask cues for success.
We thus propose TrackR-CNN as a baseline method which addresses all aspects of the MOTS task. TrackR-CNN extends Mask R-CNN \cite{He17ICCV} with 3D convolutions to incorporate temporal information and by an association head which is used to link object identities over time.

In summary, this paper makes the following \textbf{contributions:}
  (1) {We provide two new datasets with temporally consistent object instance segmentations based on the popular KITTI \cite{Geiger12CVPR} and MOTChallenge \cite{Milan16Arxiv} datasets for training and evaluating methods that tackle the MOTS task.}
  (2) {We propose the new soft Multi-Object Tracking and Segmentation Accuracy (sMOTSA) measure that can be used to simultaneously evaluate all aspects of the new task.}
  (3) {We present TrackR-CNN as a baseline method which addresses detection, tracking, and segmentation jointly and we compare it to existing work.}
  (4) {We demonstrate the usefulness of the new datasets for end-to-end training of pixel-level multi-object trackers.}
In particular, we show that with our datasets, joint training of segmentation and tracking procedures becomes possible and yields improvements over training only for instance segmentation or bounding box tracking, which was possible previously.

\section{Related Work}
\PARbegin{Multi-Object Tracking Datasets.}
In the multi-object tracking (MOT) task, an initially unknown number of targets from a known set of classes must be tracked as bounding boxes in a video. In particular, targets may enter and leave the scene at any time and must be recovered after long-time occlusion and under appearance changes. Many MOT datasets focus on street scenarios, for example the KITTI tracking dataset \cite{Geiger12CVPR}, which features video from a vehicle-mounted camera; or the MOTChallenge datasets \cite{LealTaixe15Arxiv, Milan16Arxiv} that show pedestrians from a variety of different viewpoints. UA-DETRAC \cite{Wen15Arxiv, Lyu17AVSS} also features street scenes but contains annotations for vehicles only.
Another MOT dataset is PathTrack \cite{Manen17ICCV}, which provides annotations of human trajectories in diverse scenes. PoseTrack \cite{Andriluka18CVPR} contains annotations of joint positions for multiple persons in videos.
None of these datasets provide segmentation masks for the annotated objects and thus do not describe complex interactions like in Fig.~\ref{fig:people-overlap} in sufficient detail.

\PAR{Video Object Segmentation Datasets.}
In the video object segmentation (VOS) task, instance segmentations for one or multiple generic objects
are provided in the first frame of a video and must be segmented with pixel accuracy in all subsequent frames. Existing VOS datasets contain only few objects which are also present in most frames. In addition, the common evaluation metrics for this task (region Jaccard index and boundary F-measure) do not take error cases like id switches into account that can occur when tracking multiple objects.
In contrast, MOTS focuses on a set of pre-defined classes and considers crowded scenes with many interacting objects. MOTS also adds the difficulty of discovering and tracking a varying number of new objects as they appear and disappear in a scene.

Datasets for the VOS task include the DAVIS 2016 dataset \cite{DAVIS2016}, which focuses on single-object VOS, and the DAVIS 2017 \cite{DAVIS2017} dataset, which extends the task for multi-object VOS. Furthermore, the YouTube-VOS dataset \cite{Xu18ECCV} is available and orders of magnitude larger than DAVIS. In addition, the Segtrackv2 \cite{Segtrackv2} dataset, FBMS \cite{Ochs14TPAMI} and an annotated subset of the YouTube-Objects dataset \cite{YoutubeObjectsOriginal, YoutubeObjectsSegmentation} can be used to evaluate this task.

\PAR{Video Instance Segmentation Datasets.}
Cityscapes \cite{Cordts16CVPR}, BDD \cite{Yu18Arxiv}, and ApolloScape \cite{ApolloScape} provide video data for an automotive scenario. Instance annotations, however, are only provided for a small subset of non-adjacent frames or, in the case of ApolloScape, for each frame but without object identities over time.
Thus, they cannot be used for end-to-end training of pixel-level tracking approaches.

\PAR{Methods.}
While a comprehensive review of methods proposed for the MOT or VOS tasks is outside the scope of this paper (for the former, see \eg \cite{leal2017tracking}), we will review some works that have tackled (subsets of) the MOTS task or are in other ways related to TrackR-CNN. %

Seguin \etal~\cite{seguin2016instance} derive instance segmentations from given bounding box tracks using clustering on a superpixel level, but they do not address the detection or tracking problem.
Milan \etal~\cite{Milan15CVPR} consider tracking and segmentation jointly in a CRF utilizing superpixel information and given object detections. %
In contrast to both methods, our proposed baseline operates on pixel rather than superpixel level.
CAMOT \cite{Osep18ICRA} performs mask-based tracking of generic objects on the KITTI dataset using stereo information, which limits its accuracy for distant objects.
CDTS \cite{Koh17ICCV} performs unsupervised VOS, \ie, without using first-frame information. It considers only short video clips with few object appearances and disappearances. In MOTS, however, many objects frequently enter or leave a crowded scene. %
While the above mentioned methods are able to produce tracking outputs with segmentation masks, their performance could not be evaluated comprehensively, since no dataset with MOTS annotations existed.

Lu \etal~\cite{Lu17ICCV} tackle tracking by aggregating location and appearance features per frame and combining these across time using LSTMs. Sadeghian \etal~\cite{Sadeghian17ICCV} also combine appearance features obtained by cropped detections with velocity and interaction information using a combination of LSTMs.
In both cases, the combined features are input into a traditional Hungarian matching procedure. For our baseline model, we directly enrich detections using temporal information and learn association features jointly with the detector rather than only ``post-processing'' given detections. %

\PAR{Semi-Automatic Annotation.}
There are many methods for semi-automatic instance segmentation, \eg generating segmentation masks from scribbles \cite{Rother04Siggraph}, or clicks \cite{Xu16CVPR}.
These methods require user input for every object to be segmented, while our annotation procedure can segment many objects fully-automatically, letting annotators focus on improving results for difficult cases. %
While this is somewhat similar to an active learning setting \cite{collins2008towards,vondrick2011video}, we leave the decision which objects to annotate with our human annotators to guarantee that all annotations reach the quality necessary for a long-term benchmark dataset (\cf \cite{lowell2018transferable}).

Other semi-automatic annotation techniques include Polygon-RNN \cite{Castrejon17CVPR,Acuna18CVPR}, which automatically predicts a segmentation in form of a polygon from which vertices can be corrected by the annotator. Fluid Annotation \cite{Andriluka18Arxiv} allows the annotator to manipulate segments predicted by Mask R-CNN \cite{He17ICCV} in order to annotate full images.
While speeding up the creation of segmentation masks of objects in isolated frames, these methods do not operate on a track level, do not make use of existing bounding box annotations, and do not exploit segmentation masks which have been annotated for the same object in other video frames.

\section{Datasets}
\label{sec:datasets}
Annotating pixel masks for every frame of every object in a video is an extremely time-consuming task. Hence, the availability of such data is very limited. We are not aware of any existing datasets for the MOTS task. However, there are some datasets with MOT annotations, \ie, tracks annotated at the bounding box level. For the MOTS task, these datasets lack segmentation masks. Our annotation procedure therefore adds segmentation masks for the bounding boxes in two MOT datasets. In total, we annotated 65,213 segmentation masks. This size makes our datasets viable for training and evaluating modern learning-based techniques.

\PAR{Semi-automatic Annotation Procedure.}
In order to keep the annotation effort manageable, we propose a semi-automatic method to extend bounding box level annotations by segmentation masks. 
We use a convolutional network to automatically produce segmentation masks from bounding boxes, followed by a correction step using manual polygon annotations. Per track, we fine-tune the initial network using the manual annotations as additional training data, similarly to \cite{OSVOS}. We iterate the process of generating and correcting masks until pixel-level accuracy for all annotation masks has been reached.

For converting bounding boxes into segmentation masks, we use a fully-convolutional refinement network \cite{Luiten18ACCV} %
based on DeepLabv3+ \cite{Chen18ECCV} which takes as input a crop of the input image specified by the bounding box with a small context region added, together with an additional input channel that encodes the bounding box as a mask. Based on these cues, the refinement network predicts a segmentation mask for the given box. The refinement network is pre-trained on COCO \cite{coco} and Mapillary \cite{neuhold2017mapillary}, and then trained on manually created segmentation masks for the target dataset.

In the beginning, we annotate (as polygons) two segmentation masks per object in the considered dataset.\footnote{The two frames annotated per object are chosen by the annotator based on diversity.} The refinement network is first trained on all manually created masks and afterwards fine-tuned individually for each object. These fine-tuned variants of the network are then used to generate segmentation masks for all bounding boxes of the respective object in the dataset. 
This way the network adapts to the appearance and context of each individual object.
Using two manually annotated segmentation masks per object for fine-tuning the refinement network already produces relatively good masks for the object's appearances in the other frames, but often small errors remain. Hence, we manually correct some of the flawed generated masks and re-run the training procedure in an iterative process.
Our annotators also corrected imprecise or wrong bounding box annotations in the original MOT datasets.

\begin{figure}[t]
\includegraphics[width=0.48\textwidth]{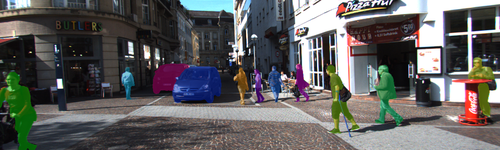}
\vspace{2pt}
\includegraphics[width=0.48\textwidth]{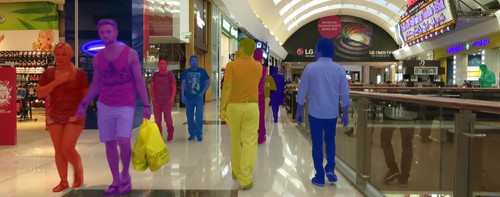}
\caption{\label{fig:kitti-and-mot-samples} \textbf{Sample Images of our Annotations.} KITTI MOTS (top) and MOTSChallenge (bottom).}
\end{figure}

\begin{table}
\small
\setlength{\tabcolsep}{3.5pt}
\centering{}%
\begin{tabular}{lccc}
\toprule 
 & \multicolumn{2}{c}{{\footnotesize{}KITTI MOTS}} & \multicolumn{1}{c}{{\footnotesize{}MOTSChallenge}}\tabularnewline
 & {\footnotesize{}train } & {\footnotesize{}val } & \tabularnewline
\midrule 
{\footnotesize{}\# Sequences } & {\footnotesize{}12 } & {\footnotesize{}9 } & {\footnotesize{}4}\tabularnewline
{\footnotesize{}\# Frames } & {\footnotesize{}5,027 } & {\footnotesize{}2,981 } & {\footnotesize{}2,862}\tabularnewline
\midrule 
{\footnotesize{}\# Tracks Pedestrian } & {\footnotesize{}99 } & {\footnotesize{}68 } & {\footnotesize{}228 }\tabularnewline
{\footnotesize{}\# Masks Pedestrian } &  &  & \tabularnewline
{\footnotesize{}\ \ \ \ \ \  Total } & {\footnotesize{}8,073 } & {\footnotesize{}3,347 } & {\footnotesize{}26,894 }\tabularnewline
{\footnotesize{}\ \ \ \ \ \  Manually annotated } & {\footnotesize{}1,312 } & {\footnotesize{}647 } & {\footnotesize{}3,930 }\tabularnewline
\midrule 
{\footnotesize{}\# Tracks Car } & {\footnotesize{}431 } & {\footnotesize{}151 } & {\footnotesize{}-}\tabularnewline
{\footnotesize{}\# Masks Car } &  &  & \tabularnewline
{\footnotesize{}\ \ \ \ \ \  Total } & {\footnotesize{}18,831 } & {\footnotesize{}8,068 } & {\footnotesize{}-}\tabularnewline
{\footnotesize{}\ \ \ \ \ \  Manually annotated } & {\footnotesize{}1,509 } & {\footnotesize{}593 } & {\footnotesize{}-}\tabularnewline
\bottomrule
\end{tabular}\caption{\label{tab:dataset-stats}\textbf{Statistics of the Introduced KITTI MOTS and MOTSChallenge
Datasets}. We consider pedestrians for both datasets and also cars for KITTI MOTS.}
\end{table}

\PAR{KITTI MOTS.}
We performed the aforementioned annotation procedure on the bounding box level annotations from the KITTI tracking dataset \cite{Geiger12CVPR}. A sample of the annotations is shown in Fig.~\ref{fig:kitti-and-mot-samples}.
To facilitate training and evaluation, we divided the 21 training sequences of the KITTI tracking dataset\footnote{We are currently applying our annotation procedure to the KITTI test set with the goal of creating a publicly accessible MOTS benchmark.} into a training and validation set, respectively\footnote{Sequences 2, 6, 7, 8, 10, 13, 14, 16 and 18 were chosen for the validation set, the remaining sequences for the training set.}. Our split balances the number of occurrences of each class -- cars and pedestrians -- roughly equally across training and validation set. Statistics are given in Table~\ref{tab:dataset-stats}. The relatively high number of manual annotations required demonstrates that existing single-image instance segmentation techniques still perform poorly on this task. This is a major motivation for our proposed MOTS dataset which allows for incorporating temporal reasoning into instance segmentation models.

\PAR{MOTSChallenge.}
We further annotated 4 of 7 sequences of the MOTChallenge 2017 \cite{Milan16Arxiv} training dataset\footnote{Sequences 2, 5, 9 and 11 were annotated.} %
and obtained the MOTSChallenge dataset.
MOTSChallenge focuses on pedestrians in crowded scenes and is very challenging due to many occlusion cases, for which a pixel-wise description is especially beneficial. A sample of the annotations is shown in Fig.~\ref{fig:kitti-and-mot-samples}, statistics are given in Table \ref{tab:dataset-stats}.

\section{Evaluation Measures}\label{sec:eval_measures}
As evaluation measures we adapt the well-established CLEAR MOT metrics for multi-object tracking \cite{Bernardin2008Eurasip} to our task. For the MOTS task, the segmentation masks per object need to be accommodated in the evaluation metric. Inspired by the Panoptic Segmentation task \cite{Kirillov18Arxiv}, we require that both the ground truth masks of objects and the masks produced by a MOTS method are non-overlapping, \ie, each pixel can be assigned to at most one object. We now introduce our evaluation measures for MOTS.

Formally, the ground truth of a video with $T$ time frames, height $h$, and width $w$ consists of a set of $N$ non-empty ground truth pixel masks $M=\{m_1, \dots, m_N\}$ with $m_i \in \{0,1\}^{h \times w}$, each of which belongs to a corresponding time frame $t_m \in\{1,\dots,T\}$ and is assigned a ground truth track id $id_m \in \mathbb{N}$. The output of a MOTS method is a set of $K$ non-empty hypothesis masks $H=\{h_1, \dots, h_K\}$ with $h_i \in \{0,1\}^{h \times w}$, each of which is assigned a hypothesized track id $id_h \in \mathbb{N}$ and a time frame $t_h \in \{1, \dots, T\}$.

\PAR{Establishing Correspondences.}
An important step for the CLEAR MOT metrics \cite{Bernardin2008Eurasip} is to establish correspondences between ground truth objects and tracker hypotheses. In the bounding box-based setup, establishing correspondences is non-trivial and performed by bipartite matching, since ground truth boxes may overlap and multiple hypothesized boxes can fit well to a given ground truth box.
In the case of MOTS, establishing correspondences is greatly simplified since we require that each pixel is uniquely assigned to at most one object in the ground truth and the hypotheses respectively. Thus, at most one predicted mask can have an Intersection-over-Union (IoU) of more than $0.5$ with a given ground truth mask \cite{Kirillov18Arxiv}. Hence, the mapping $c:\ H\rightarrow M\cup\{\emptyset\}$ from hypothesis masks to ground truth masks can simply be defined using mask-based IoU as
\vspace{-6pt}
\begin{eqnarray}
\small
c(h)=\begin{cases}
\displaystyle{\mathop{\arg\max}_{m\in M}}\ \text{IoU}(h,m), & \text{if}\ \displaystyle{\max_{m\in M}}\ \text{IoU}(h,m)>0.5\\
\emptyset, & \text{otherwise}.
\end{cases}
\\[-16pt]\nonumber
\end{eqnarray}

The set of true positives $\text{TP}=\{h\in H\mid c(h)\neq\emptyset\}$ is comprised of hypothesized masks which are mapped to a ground truth mask. Similarly, false positives are hypothesized masks that are not mapped to any ground truth mask, \ie $\text{FP}=\{h\in H\mid c(h)=\emptyset\}$.
Finally, the set $\text{FN}=\{m \in M\mid c^{-1}(m)=\emptyset\}$ of false negatives contains the ground truth masks which are not covered by any hypothesized mask.

\begin{figure*}[t] %
\includegraphics[width=1.0\textwidth]{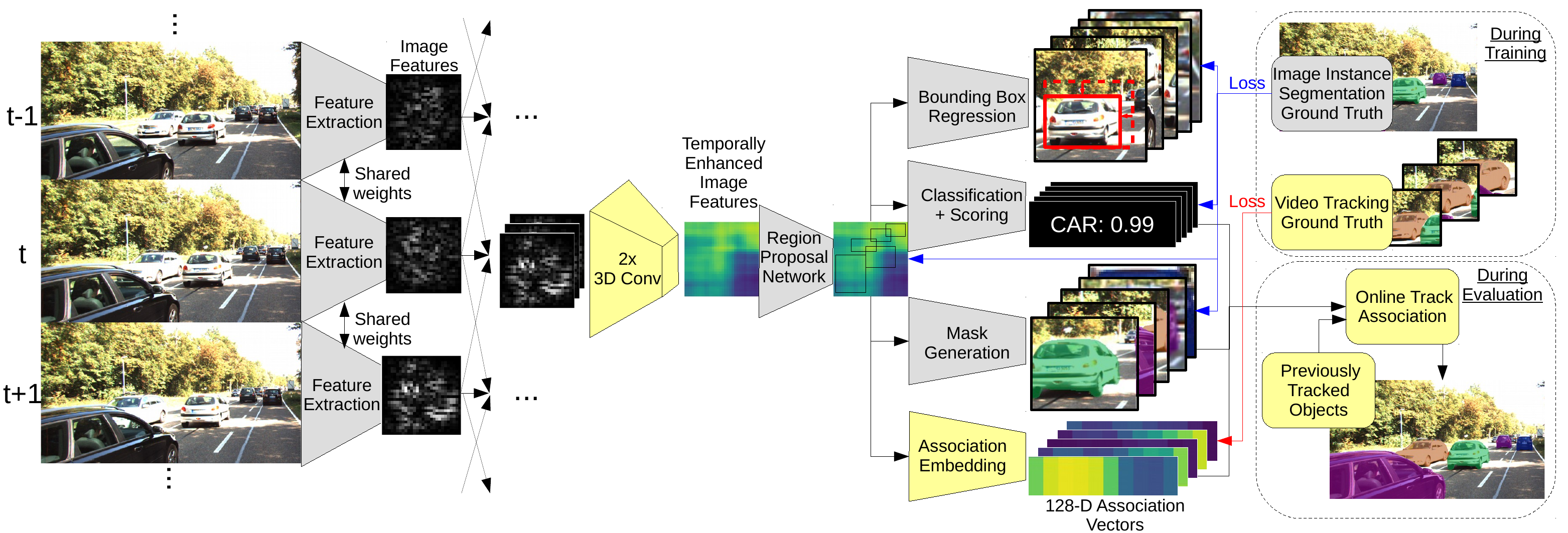}
\caption{\label{fig:overview} \textbf{TrackR-CNN Overview.} We extend Mask R-CNN by 3D convolutions to incorporate temporal context and by an association head that produces association vectors for each detection. The Euclidean distances between association vectors are used to associate detections over time into tracks. Differences to Mask R-CNN are highlighted in yellow.}
\end{figure*}

In the following, let $\mathit{pred}: M \rightarrow M \cup \{\emptyset\}$ denote the latest tracked predecessor of a ground truth mask, or $\emptyset$ if no tracked predecessor exists. So $q=\mathit{pred}(p)$ is the mask $q$ with the same id ($id_q=id_p$) and the largest $t_q<t_p$ such that $c^{-1}(q)\neq\emptyset$ \footnote{This definition corresponds to the one used by MOTChallenge. Note that the original KITTI tracking benchmark does not count id switches if the target was lost by the tracker.}.
The set $\text{IDS}$ of id switches is then defined as the set of ground truth masks whose predecessor was tracked with a different id. Formally,
\begin{eqnarray}
\small
\begin{split}
  \text{IDS}=\{&m\in M\mid c^{-1}(m)\neq\emptyset\land \mathit{pred}(m)\neq\emptyset\ \land\\& id_{c^{-1}(m)}\neq id_{c^{-1}(\mathit{pred}(m))}\} .
\end{split}
\end{eqnarray}

\PAR{Mask-based Evaluation Measures.}
Additionally, we define a soft version $\widetilde{\text{TP}}$ of the number of true positives by
\vspace{-6pt}
\begin{eqnarray}
  \widetilde{\text{TP}}=\sum_{h\in TP}\text{IoU}(h,c(h)).
\\[-16pt]\nonumber
\end{eqnarray}
Given the previous definitions, we define mask-based variants of the original CLEAR MOT metrics \cite{Bernardin2008Eurasip}.
We propose the multi-object tracking and segmentation accuracy (MOTSA) as a mask IoU based version of the box-based MOTA metric, \ie
\begin{equation}
\scriptsize
\text{MOTSA}=1-\frac{|FN|+|FP|+|IDS|}{|M|}%
    =\frac{|TP|-|FP|-|IDS|}{|M|},
\end{equation}
and the mask-based multi-object tracking and segmentation precision (MOTSP) as
\vspace{-12pt}
\begin{eqnarray}
\text{MOTSP}=\frac{\widetilde{TP}}{|TP|}.
\\[-16pt]\nonumber
\end{eqnarray}
Finally, we introduce the soft multi-object tracking and segmentation accuracy (sMOTSA)
\vspace{-6pt}
\begin{eqnarray}
\text{sMOTSA}=\frac{\widetilde{TP}-|FP|-|IDS|}{|M|},
\\[-16pt]\nonumber
\end{eqnarray}
which accumulates the soft number $\widetilde{\text{TP}}$ of true positives instead of counting how many masks reach an IoU of more than $0.5$.~sMOTSA therefore measures segmentation as well as detection and tracking quality.

\section{Method}
\label{sec:method}
In order to tackle detection, tracking, and segmentation, \ie the MOTS task, jointly %
with a neural network, we build upon the popular Mask R-CNN \cite{He17ICCV} architecture, which extends the Faster R-CNN \cite{Ren15NIPS} detector with a mask head. We propose TrackR-CNN (see Fig.~\ref{fig:overview}) which in turn extends Mask R-CNN by an association head and two 3D convolutional layers to be able to associate detections over time and deal with temporal dynamics. TrackR-CNN provides mask-based detections together with association features. Both are input to a tracking algorithm that decides which detections to select and how to link them over time.

\PAR{Integrating temporal context.} In order to exploit the temporal context of the input video \cite{Carreira17CVPR}, we integrate 3D convolutions (where the additional third dimension is time) into Mask R-CNN on top of a ResNet-101 \cite{resnet} backbone.
The 3D convolutions are applied to the backbone features in order to augment them with temporal context. These augmented features are then used by the region proposal network (RPN).
As an alternative we also consider convolutional LSTM \cite{NIPS15Shi,Liu18CVPR} layers. Convolutional LSTM retains the spatial structure of the input by calculating its activations using convolutions instead of matrix products.

\PAR{Association Head.} In order to be able to associate detections over time, we extend Mask R-CNN by an association head which is a fully connected layer that gets region proposals as inputs and predicts an association vector for each proposal. The association head is inspired by the embedding vectors used in person re-identification \cite{Hermans17Arxiv, Beyer17CVPRW, Long18ICME, Siyu17CVPR, Zheng17TOMM}.
Each association vector represents the identity of a car or a person. They are trained in a way that vectors belonging to the same instance are close to each other and vectors belonging to different instances are far away from each other. We define the distance $d(v,w)$ between two association vectors $v$ and $w$ as their Euclidean distance, \ie
\vspace{-10pt}
\begin{eqnarray}
  d(v,w):=\lVert v- w \rVert.
  \label{eq:assoc-dist}
\\[-18pt]\nonumber
\end{eqnarray}

We train the association head using the batch hard triplet loss proposed by Hermans \etal \cite{Hermans17Arxiv} adapted for video sequences.
This loss samples hard positives and hard negatives for each detection. Formally,
let $\mathcal{D}$ denote the set of detections for a video. Each detection $d\in \mathcal{D}$ consists of a mask $mask_d$ and an association vector $a_d$, which come from time frame $t_d$, and is assigned a ground truth track id $id_d$ determined by its overlap with the ground truth objects.
For a video sequence of $T$ time steps, the association loss in the batch-hard formulation with margin $\alpha$ is  then given by
\begin{equation}\label{eq:batch_hard_loss}
\frac{1}{|D|}\sum_{d\in\mathcal{D}}\max\big(\max_{\stackrel{e\in\mathcal{D}:}{id_{e}=id_{d}}}\lVert a_{e}-a_{d}\rVert-\min_{\stackrel{e\in\mathcal{D}:}{id_{e}\neq id_{d}}}\lVert a_{e}-a_{d}\rVert+\alpha,0\big).
\end{equation}

\PAR{Mask Propagation.}
Mask-based IoU together with optical flow warping is a strong cue for associating pixel masks over time \cite{Osep18ICRA, Luiten18ACCV}.
Hence, we also experiment with %
mask warping as an alternative cue to association vector similarities. For a detection $d\in \mathcal{D}$ at time $t-1$ with mask $mask_d$ and a detection $e\in \mathcal{D}$ with mask $mask_e$ at time $t$, we define the mask propagation score \cite{Luiten18ACCV,Osep18ICRA} as
\begin{equation}
\small
  \text{maskprop}(mask_d, mask_e)=\text{IoU}(\mathcal{W}(mask_d),   mask_e),
  \label{eq:maskprop}
\end{equation}
where $\mathcal{W}(m)$ denotes warping mask $m$ forward by the optical flow between frames $t-1$ and $t$.

\PAR{Tracking.}\label{par:tracking}
In order to produce the final result, we still need to decide which detections to report and how to link them into tracks over time. For this, we extend existing tracks with new detections based on their association vector similarity to the most recent detection in that track.

More precisely, for each class and each frame $t$, we link together detections at the current frame that have detector confidence larger than a threshold $\gamma$  with detections selected in the previous frames using the association vector distances from Eq.~\ref{eq:assoc-dist}. We only choose the most recent detection for tracks from up to a threshold of $\beta$ frames in the past.
Matching is done with the Hungarian algorithm, while only allowing pairs of detections with a distance smaller than a threshold $\delta$.
Finally, all unassigned high confidence detections start new tracks.

The resulting tracks can contain overlapping masks which we do not allow for the MOTS task (\cf Section~\ref{sec:eval_measures}). In such a case, pixels belonging to detections with a higher confidence (given by the classification head of our network) take precedence over detections with lower confidence.

\section{Experiments}
\PARbegin{Experimental Setup.}
For Mask R-CNN we use a ResNet-101 backbone \cite{resnet} and pre-train it on COCO \cite{coco} and Mapillary \cite{neuhold2017mapillary}. Afterwards, we construct TrackR-CNN by adding the association head and integrating two depthwise separable 3D convolution layers with $3\times3\times3$ filter kernels each (two dimensions are spatial and the third is over time), ReLU activation, and $1024$ feature maps between the backbone and the region proposal network. The 3D convolutions are initialized to an identity function after which the ReLU is applied. When using convolutional LSTM, weights are initialized randomly and a skip connection is added to preserve activations for the pretrained weights of subsequent layers during the initial steps of training.
TrackR-CNN is then trained on the target dataset, \ie KITTI MOTS or MOTSChallenge, for 40 epochs with a learning rate of $5\cdot10^{-7}$ using the Adam \cite{adam} optimizer. During training, mini-batches which consist of $8$ adjacent frames of a single video are used, where $8$ was the maximum possible number of frames to fit into memory with a Titan X (Pascal) graphics card. At batch boundaries, the input to the 3D convolution layer is zero padded in time. When using convolutional LSTM, gradients are backpropagated through all 8 frames during training and at test time the recurrent state is propagated over the whole sequence.
The vectors produced by the association head have 128 dimensions and the association loss defined in Eq.~\ref{eq:batch_hard_loss} is computed over the detections obtained in one batch. We choose a margin of $\alpha=0.2$, which proved useful in \cite{Hermans17Arxiv}.
For the mask propagation experiments, we compute optical flow between all pairs of adjacent frames using PWC-Net \cite{Sun18CVPR}.
Our whole tracker achieves a speed of around 2 frames per second at test time. When using convolutional LSTM, it runs online and when using 3D convolutions in near-online fashion due to the two frames look-ahead of the 3D convolutions.

We tune the thresholds for our tracking system ($\delta$,  $\beta$, $\gamma$) for each class separately on the target training set with random search using $1000$ iterations per experiment.

\begin{table}[t]
\setlength{\tabcolsep}{3.5pt}
\begin{centering}
\begin{tabular}{ccccccc}
\toprule
\multirow{2}{*}{} & \multicolumn{2}{c}{{\scriptsize{}sMOTSA}} & \multicolumn{2}{c}{{\scriptsize{}MOTSA}} & \multicolumn{2}{c}{{\scriptsize{}MOTSP}}\tabularnewline
 & {\scriptsize{}Car} & {\scriptsize{}Ped} & {\scriptsize{}Car} & {\scriptsize{}Ped} & {\scriptsize{}Car} & {\scriptsize{}Ped}\tabularnewline
\midrule
{\footnotesize{}TrackR-CNN (ours)} & {\footnotesize{}76.2} & {\footnotesize{}46.8} & {\footnotesize{}87.8} & {\footnotesize{}65.1} & {\footnotesize{}\textbf{87.2}} & {\footnotesize{}\textbf{75.7}}\tabularnewline
{\footnotesize{}Mask R-CNN + maskprop} & {\footnotesize{}75.1} & {\footnotesize{}45.0} & {\footnotesize{}86.6} & {\footnotesize{}63.5} & {\footnotesize{}87.1} & {\footnotesize{}75.6}\tabularnewline
{\footnotesize{}TrackR-CNN (box orig) + MG}& {\footnotesize{}75.0} & {\footnotesize{}41.2} & {\footnotesize{}87.0} & {\footnotesize{}57.9} & {\footnotesize{}86.8} & {\footnotesize{}76.3}\tabularnewline
{\footnotesize{}TrackR-CNN (ours) + MG} & {\footnotesize{}76.2} & {\footnotesize{}\textbf{47.1}} & {\footnotesize{}87.8} & {\footnotesize{}\textbf{65.5}} & {\footnotesize{}\textbf{87.2}} & {\footnotesize{}\textbf{75.7}}\tabularnewline
\midrule
{\footnotesize{}CAMOT \cite{Osep18ICRA} (our det)} & {\footnotesize{}67.4} & {\footnotesize{}39.5} & {\footnotesize{}78.6} & {\footnotesize{}57.6} & {\footnotesize{}86.5} & {\footnotesize{}73.1}\tabularnewline
{\footnotesize{}CIWT \cite{Osep17ICRA} (our det) + MG} & {\footnotesize{}68.1} & {\footnotesize{}42.9} & {\footnotesize{}79.4} & {\footnotesize{}61.0} & {\footnotesize{}86.7} & {\footnotesize{}75.7}\tabularnewline
\midrule
{\footnotesize{}BeyondPixels \cite{SharmaICRA18} + MG} & {\footnotesize{}\textbf{76.9}} & {\footnotesize{}-} & {\footnotesize{}\textbf{89.7}} & {\footnotesize{}-} & {\footnotesize{}86.5} & {\footnotesize{}-}\tabularnewline
\midrule
\midrule
{\footnotesize{}GT Boxes (orig) + MG} & {\footnotesize{}77.3} & {\footnotesize{36.5}} & {\footnotesize{}90.4} & {\footnotesize{}55.7} & {\footnotesize{}86.3} & {\footnotesize{}75.3}\tabularnewline
{\footnotesize{}GT Boxes (tight) + MG} & {\footnotesize{} \textbf{82.5}} & {\footnotesize{} \textbf{50.0}} & {\footnotesize{} \textbf{95.3}} & {\footnotesize{} \textbf{71.1}} & {\footnotesize{} \textbf{86.9}} & {\footnotesize{} \textbf{75.4}}\tabularnewline
\bottomrule
\end{tabular}
\par\end{centering}
\caption{\label{tab:results-kitti-main}\textbf{Results on KITTI MOTS}. 
+MG denotes mask generation with a KITTI MOTS fine-tuned Mask R-CNN. BeyondPixels is a state-of-the-art MOT method for cars and uses a different detector than the other methods.}
\end{table}

\newcommand\img[2]%
  {\raisebox
    {\dimexpr-0.5\height+0.5ex}%
    {\includegraphics[width=#2]{#1}}%
  }
\begin{figure}[t]
  \begin{imgrows}[\textwidth]
    \newcommand{\mysize}{26mm}
    \imgrow
      \img{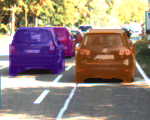}{\mysize}
      \hspace{-4pt}
      \vspace{0.5pt}
      \img{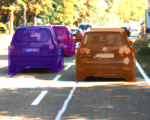}{\mysize}
      \hspace{-4pt}
      \img{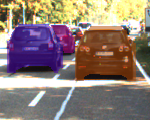}{\mysize}
    \imgrow
      \img{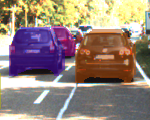}{\mysize}
      \hspace{-4pt}
      \vspace{0.5pt}
      \img{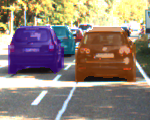}{\mysize}
      \hspace{-4pt}
      \img{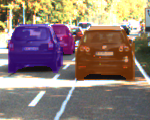}{\mysize}
    \imgrow
      \img{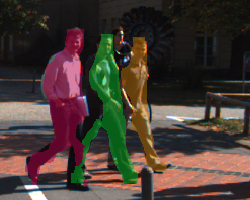}{\mysize}
      \vspace{0.5pt}
      \hspace{-4pt}
      \img{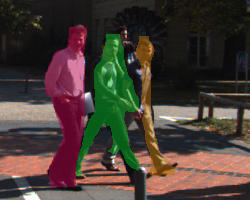}{\mysize}
      \hspace{-4pt}
      \img{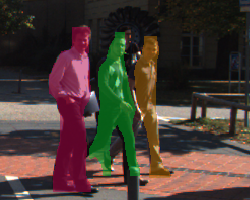}{\mysize}
    \imgrow
      \img{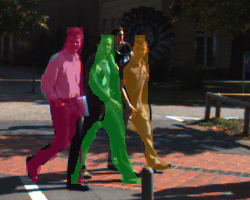}{\mysize}
      \hspace{-4pt}
      \img{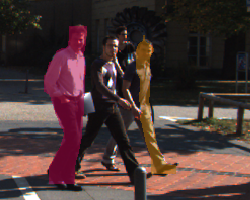}{\mysize}
      \hspace{-4pt}
      \img{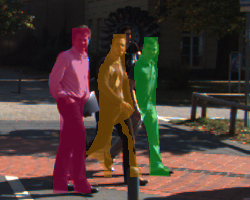}{\mysize}
  \end{imgrows}
\caption{\textbf{Qualitative Results on KITTI MOTS.} (a)+(c) Our TrackR-CNN model evaluated on validation sequences of KITTI MOTS. (b)+(d) TrackR-CNN (box orig) + MG evaluated on the same sequences. Training with masks on our data avoids confusion between similar near-by objects.}
\label{fig:results_example}
\end{figure}

\PAR{Main Results.}
Table~\ref{tab:results-kitti-main} shows our results on the KITTI MOTS validation set. %
We achieve competitive results, beating several baselines.
\textit{Mask R-CNN + maskprop} denotes a simple baseline for which we fine-tuned the COCO and Mapillary pre-trained Mask R-CNN on the frames of the KITTI MOTS training set. We then evaluated it on the validation set and linked the mask-based detections over time using mask propagation scores (\cf Section~\ref{par:tracking}). Compared to this baseline, TrackR-CNN achieves higher sMOTSA and MOTSA scores, implying that the 3D convolution layers and the association head help with identifying objects in video. MOTSP scores remain similar.

\textit{TrackR-CNN (box orig)} denotes a version of our model trained without mask head on the original bounding box annotations of KITTI. We then tuned for MOTA scores according to the original KITTI tracking annotations on our training split. We evaluate this baseline in our MOTS setting by adding segmentation masks as a post-processing step (denoted by +MG) with the mask head of the KITTI fine-tuned Mask R-CNN. sMOTSA and MOTSA scores for this setup are worse than for our method and the previous baseline, especially when considering pedestrians, adding to our observation that non-tight bounding boxes are not an ideal cue for tracking and that simply using an instance segmentation method on top of bounding box predictions is not sufficient to solve the MOTS task.
We show qualitative results for this baseline in Figure~\ref{fig:results_example}.  The box-based model often confuses similar occluding objects for one another, leading to missed masks and id switches. In contrast, our model hypothesizes consistent masks.

To show that adding segmentation masks as done above does not give an unfair (dis)advantage, we also use the Mask R-CNN mask head to replace the masks generated by our method (\textit{TrackR-CNN (ours) + MG}). The results stay roughly similar, so no major (dis)advantage incurs.

In combination, our baseline experiments show that training on temporally consistent instance segmentation data for video gives advantages both over training on instance segmentation data without temporal information and over training just on bounding box tracking data. Joint training on both was not possible before, which underlines the usefulness of our proposed MOTS datasets.

\textit{CAMOT} \cite{Osep18ICRA} is a mask-based tracker which can track both objects from pre-defined classes and generic objects using 3D information from the stereo setup in KITTI. In the original version, \textit{CAMOT} takes as input generic object proposals from SharpMask \cite{Pinheiro16ECCV}. For better comparability, we used the detections from our TrackR-CNN (obtained by running it as a normal detector without association) as inputs instead. Note that \textit{CAMOT} can only track regions for which depth from stereo is available which limits its recall. The results show that our proposed tracking method performs significantly better than \textit{CAMOT} when using the same set of input detections.

Since there are not many mask-based trackers with source code available, we also considered the bounding box-based tracking methods \textit{CIWT} \cite{Osep17ICRA} and \textit{BeyondPixels} \cite{SharmaICRA18} and again converted their results to segmentation masks using the KITTI fine-tuned Mask R-CNN mask head. Note that these methods were tuned to perform well on the original bounding box based task.

\textit{CIWT} \cite{Osep17ICRA} combines image-based information with 3D information from stereo for tracking jointly in image and world space. Once more, detections from our TrackR-CNN were used for comparability. Our proposed tracking system which tackles tracking and mask generation jointly performs better than \textit{CIWT} when generating masks post-hoc.

\textit{BeyondPixels} \cite{SharmaICRA18} is one of the strongest tracking methods for cars on the original KITTI tracking dataset. It combines appearance information with 3D cues. We were not able to run their method with our detections since their code for extracting appearance features is not available. Instead we used their original detections which are obtained from RRC \cite{Ren17CVPR}, a very strong detector. RRC achieves precise localization on KITTI in particular, while the more  conventional Mask R-CNN detector was designed for general object detection. The resulting sMOTSA and MOTSA scores are higher than for our method, but still show that there are limits to state-of-the-art bounding box tracking methods on MOTS when simply segmenting boxes using Mask R-CNN.

\begin{table}
\setlength{\tabcolsep}{5pt}
\begin{centering}
\begin{tabular}{ccccccc}
\toprule
\multirow{2}{*}{{\footnotesize{}Temporal component}} & \multicolumn{2}{c}{{\footnotesize{}sMOTSA}} & \multicolumn{2}{c}{{\footnotesize{}MOTSA}} & \multicolumn{2}{c}{{\footnotesize{}MOTSP}}\tabularnewline
 & {\footnotesize{}Car} & {\footnotesize{}Ped} & {\footnotesize{}Car} & {\footnotesize{}Ped} & {\footnotesize{}Car} & {\footnotesize{}Ped}\tabularnewline
\midrule
{\footnotesize{}1xConv3D} & {\footnotesize{}76.1} & {\footnotesize{}46.3} & {\footnotesize{}87.8} & {\footnotesize{}64.5} & {\footnotesize{}87.1} & {\footnotesize{}\textbf{75.7}}\tabularnewline
{\footnotesize{}2xConv3D} & {\footnotesize{}76.2} & {\footnotesize{}\textbf{46.8}} & {\footnotesize{}87.8} & {\footnotesize{}\textbf{65.1}} & {\footnotesize{}87.2} & {\footnotesize{}\textbf{75.7}}\tabularnewline
{\footnotesize{}1xConvLSTM} & {\footnotesize{}75.7} & {\footnotesize{}45.0} & {\footnotesize{}87.3} & {\footnotesize{}63.4} & {\footnotesize{}87.2} & {\footnotesize{}75.6}\tabularnewline
{\footnotesize{}2xConvLSTM} & {\footnotesize{}76.1} & {\footnotesize{}44.8} & {\footnotesize{}\textbf{87.9}} & {\footnotesize{}63.3} & {\footnotesize{}87.0} & {\footnotesize{}75.2}\tabularnewline
{\footnotesize{}None} & {\footnotesize{}\textbf{76.4}} & {\footnotesize{}44.8} & {\footnotesize{}\textbf{87.9}} & {\footnotesize{}63.2} & {\footnotesize{}\textbf{87.3}} & {\footnotesize{}75.5}\tabularnewline
\bottomrule
\end{tabular}
\par\end{centering}
\caption{\label{tab:results-recurrent}\textbf{Different Temporal Components for TrackR-CNN}. Comparison of results on KITTI MOTS.}
\end{table}

\PAR{MOTS Using Ground Truth Boxes.}
For comparison, we derived segmentation results based on bounding box ground truth and evaluated it on our new annotations. %
Here, we consider two variants of the ground truth: the original bounding boxes from KITTI (\textit{orig}), which are amodal, \ie if only the upper body of a person is visible, the box will still extend to the ground, and tight bounding boxes (\textit{tight}) derived from our segmentation masks. Again, we generated masks using the KITTI MOTS fine-tuned Mask R-CNN.
Our results show that even with perfect track hypotheses generating accurate masks remains challenging, especially for pedestrians. This is even more the case when using amodal boxes, which often contain large regions that do not show the object. This further validates our claim that MOT tasks can benefit from pixel-wise evaluation. %
Further baselines, where we fill the ground truth boxes with rectangles or ellipses can be found in the supplemental material.

\PAR{Temporal Component.}
In Table~\ref{tab:results-recurrent}, we compare different variants of temporal components for TrackR-CNN. \textit{1xConv3D} and \textit{2xConv3D} means using either one or stacking two depthwise separable 3D convolutional layers between backbone and region proposal network, each with 1024 dimensions. Similarly, \textit{1xConvLSTM} and \textit{2xConvLSTM} denotes one or two stacked convolutional LSTM layers at the same stage with 128 feature channels each. The number of parameters per feature channel in a convolutional LSTM is higher due to gating. Using more feature channels did not seem to be helpful during initial experiments. Finally, \textit{None} denotes adding no additional layers as temporal component.
Compared to the \textit{None} baseline, adding two 3D convolutions significantly improves sMOTSA and MOTSA scores for pedestrians, while performance for cars remains comparable. Surprisingly, using convolutional LSTM does not yield any
significant gains over the baseline.

\begin{table}[t]
\setlength{\tabcolsep}{4.5pt}
\begin{centering}
\begin{tabular}{ccccccc}
\toprule
\multirow{2}{*}{{\footnotesize{}Association Mechanism}} & \multicolumn{2}{c}{{\footnotesize{}sMOTSA}} & \multicolumn{2}{c}{{\footnotesize{}MOTSA}} & \multicolumn{2}{c}{{\footnotesize{}MOTSP}}\tabularnewline
 & {\footnotesize{}Car} & {\footnotesize{}Ped} & {\footnotesize{}Car} & {\footnotesize{}Ped} & {\footnotesize{}Car} & {\footnotesize{}Ped}\tabularnewline
\midrule
{\footnotesize{}Association head} & {\footnotesize{}\textbf{76.2}} & {\footnotesize{}\textbf{46.8}} & {\footnotesize{}\textbf{87.8}} & {\footnotesize{}\textbf{65.1}} & {\footnotesize{}\textbf{87.2}} & {\footnotesize{}\textbf{75.7}}\tabularnewline
{\footnotesize{}Mask IoU} & {\footnotesize{}75.5} & {\footnotesize{}46.1} & {\footnotesize{}87.1} & {\footnotesize{}64.4} & {\footnotesize{}\textbf{87.2}} & {\footnotesize{}\textbf{75.7}}\tabularnewline
{\footnotesize{}Mask IoU (train w/o assoc.)} & {\footnotesize{}74.9} & {\footnotesize{}44.9} & {\footnotesize{}86.5} & {\footnotesize{}63.3} & {\footnotesize{}87.1} & {\footnotesize{}75.6}\tabularnewline
{\footnotesize{}Bbox IoU} & {\footnotesize{}75.4} & {\footnotesize{}45.9} & {\footnotesize{}87.0} & {\footnotesize{}64.3} & {\footnotesize{}\textbf{87.2}} & {\footnotesize{}\textbf{75.7}}\tabularnewline
{\footnotesize{}Bbox Center} & {\footnotesize{}74.3} & {\footnotesize{}43.3} & {\footnotesize{}86.0} & {\footnotesize{}61.7} & {\footnotesize{}\textbf{87.2}} & {\footnotesize{}\textbf{75.7}}\tabularnewline
\bottomrule
\end{tabular}
\par\end{centering}
\caption{\label{tab:results-association-mech}\textbf{Different Association Mechanisms for TrackR-CNN}. Comparison of results on KITTI MOTS.}
\end{table}

\PAR{Association Mechanism.}
In Table~\ref{tab:results-association-mech}, we compare different mechanism used for association between detections. Each line follows the proposed tracking system explained in Section~\ref{sec:method}, but different scores are used for the Hungarian matching step.
When using the association head, association vectors may match with detections up to $\beta$ frames in the past. For the remaining association mechanisms, only matching between adjacent frames is sensible.

For \textit{Mask IoU} we only use mask propagation scores from Eq.~\ref{eq:maskprop}, which degrades sMOTSA and MOTSA scores. This underlines the usefulness of our association head which can outperform an optical flow based cue using embeddings provided by a single neural network. Here, we also try training without the association loss (\textit{Mask IoU (train w/o assoc.)}), which degrades MOTSA scores even more. Therefore, the association loss also has a positive effect on the detector itself.
Surprisingly, using bounding box IoU (where the boxes were warped with the median of the optical flow values inside the box, \textit{Bbox IoU}) performs almost the same as mask IoU.
Finally, using only distances of bounding box centers (\textit{Bbox Center}) for association, \ie doing a nearest neighbor search, significantly degrades performance.

\PAR{MOTSChallenge.}
Table \ref{tab:results-mot} shows our results on the MOTSChallenge dataset. Since MOTSChallenge only has four video sequences, we trained our method (\textit{TrackR-CNN (ours)}) in a leaving-one-out fashion (evaluating each sequence with a model trained and tuned on the three others).

For comparison, we took pre-computed results of four methods that perform well on the MOT17 benchmark and generated masks using a Mask R-CNN fine-tuned on MOTSChallenge (in a leaving-one-out fashion) to evaluate them on our data. We note that all four sets of results use the strongest set of public detections generated with SDP \cite{Yang16CVPR}, while TrackR-CNN generates its own detections. It is also unclear how much these methods were trained to perform well on the MOTChallenge training set, on which MOTSChallenge is based. Despite these odds, TrackR-CNN outperforms all other methods.
The last line demonstrates that even with the tight ground truth bounding boxes including track information over time, segmenting all pedestrians accurately remains difficult.

\begin{table}[t]
\footnotesize
\setlength{\tabcolsep}{9.5pt}
\begin{centering}
\begin{tabular}{cccc}
\toprule
\multirow{1}{*}{} & \multicolumn{1}{c}{{\footnotesize{}sMOTSA}} & \multicolumn{1}{c}{{\footnotesize{}MOTSA}} & {\footnotesize{}MOTSP}\tabularnewline
\midrule
{\footnotesize{}TrackR-CNN (ours)} & \textbf{52.7} & \textbf{66.9} & \textbf{80.2} \tabularnewline
\midrule
{\footnotesize{}MHT-DAM \cite{Kim15ICCV} + MG} & 48.0 & 62.7 & 79.8 \tabularnewline
{\footnotesize{}FWT \cite{HenLea2018} + MG} & 49.3 & 64.0 & 79.7 \tabularnewline
{\footnotesize{}MOTDT \cite{Long18ICME} + MG} & 47.8 & 61.1 & 80.0 \tabularnewline
{\footnotesize{}jCC \cite{Keuper18TPAMI} + MG} & 48.3  & 63.0 & 79.9 \tabularnewline
\midrule
\midrule
{\footnotesize{}GT Boxes (tight) + MG} & \textbf{55.8} & \textbf{74.5} & \textbf{78.6} \tabularnewline
\bottomrule
\end{tabular}
\par\end{centering}
\caption{\label{tab:results-mot}\textbf{Results on MOTSChallenge}. +MG denotes mask generation with a domain fine-tuned Mask R-CNN.}
\end{table}

\section{Conclusion}

Until now there has been no benchmark or dataset to evaluate the task of  multi-object tracking and segmentation and to directly train methods using such temporally consistent mask-based tracking information.
To alleviate this problem, we introduce two new datasets based on existing MOT datasets which we annotate using a semi-automatic annotation procedure. We further introduce the MOTSA and sMOTSA metrics, based on the commonly used MOTA metric, but adapted to evaluate all aspects of mask-based tracking.
We finally develop a baseline model that was designed to take advantage of this data. We show that through training on our data, the method is able to outperform comparable methods which are only trained with bounding box tracks and single image instance segmentation masks. Our new datasets now make such joint training possible, which opens up many opportunities for future research.

\footnotesize \PAR{Acknowledgements:} This project has been funded, in parts, by ERC Consolidator Grant DeeViSe (ERC-2017-COG-773161). The experiments were performed with computing resources granted by RWTH Aachen University under project rwth0373. We would like to thank our annotators.

{\small
\bibliographystyle{ieee}
\bibliography{abbrev_short,paper}
}

\clearpage
\appendix
\normalsize
\part*{Supplemental Material}

\section{Losses for the Association Head}
TrackR-CNN uses association scores based on vectors predicted by an association head to identify the same object across time. In our baseline model, we train this head using a batch hard triplet loss proposed by Hermans \etal \cite{Hermans17Arxiv}, which we state again here:
Let $\mathcal{D}$ denote the set of detections for a video. Each detection $d\in \mathcal{D}$ has a corresponding association vector $a_d$ and is assigned a ground truth track id $id_d$ determined by its overlap with the ground truth objects (we only consider detections which sufficiently overlap with a ground truth object here).
For a video sequence of $T$ time steps, the association loss in the batch-hard formulation with margin $\alpha$ is then given by
\begin{equation}\label{eq:batch_hard_loss_supp}
\begin{split}
\mathcal{L}_{batch\_hard} = \frac{1}{|D|}\sum_{d\in\mathcal{D}}\max\big(\max_{\stackrel{e\in\mathcal{D}:}{id_{e}=id_{d}}}\lVert a_{e}-a_{d}\rVert-\\
\min_{\stackrel{e\in\mathcal{D}:}{id_{e}\neq id_{d}}}\lVert a_{e}-a_{d}\rVert+\alpha,0\big).
\end{split}
\end{equation}
Intuitively, each detection $d$ is selected as an anchor and then the most dissimilar detection with the same id is selected as a hard positive example and the most similar detection with a different id is selected as a hard negative example for this anchor. The margin $\alpha$ and maximum operation ensure that the distance of the anchor to the hard positive is smaller than its distance to the hard negative example by at least $\alpha$.

In order to justify our choice of the batch-hard loss, we also report results using two alternative loss formulations, namely the batch all loss \cite{Hermans17Arxiv} which considers all pairs of detections, \ie
\begin{equation}
\begin{split}
\mathcal{L}_{batch\_all}=\frac{1}{|\mathcal{D}|^{2}}\sum_{d\in\mathcal{D}}\sum_{e\in D}\max\big(\lVert a_{e}-a_{d}\rVert-\\
\lVert a_{e}-a_{d}\rVert+\alpha,0\big)
\end{split}
\end{equation}
and the contrastive loss \cite{Hadsell06CVPR}
\begin{equation}
\begin{split}
\mathcal{L}_{contrastive}=
\frac{1}{|\mathcal{D}|^{2}}\big(&\sum_{d\in\mathcal{D}}\sum_{\stackrel{e\in\mathcal{D}}{id_{e}=id_{d}}}\lVert a_{e}-a_{d}\rVert^{2}+\\&\sum_{d\in\mathcal{D}}\sum_{\stackrel{e\in\mathcal{D}}{id_{e}\neq id_{d}}}
\max(\alpha-\lVert a_{e}-a_{d}\rVert,0)^{2}\big).
\end{split}
\end{equation}

Table \ref{tab:results-embedding} compares the performance of these different variants of the loss function on the KITTI MOTS validation set. It can be seen that the batch hard triplet loss performs better than just considering all pairs of detections (\textit{Batch All Triplet}), or using the conventional contrastive loss (\textit{Contrastive}). Especially for pedestrians performance using the contrastive loss is low.

\begin{table}[h]
\begin{centering}
\begin{tabular}{ccccccc}
\toprule
\multirow{2}{*}{{\footnotesize{}Association Loss}} & \multicolumn{2}{c}{{\footnotesize{}sMOTSA}} & \multicolumn{2}{c}{{\footnotesize{}MOTSA}} & \multicolumn{2}{c}{{\footnotesize{}MOTSP}}\tabularnewline
 & {\footnotesize{}Car} & {\footnotesize{}Ped} & {\footnotesize{}Car} & {\footnotesize{}Ped} & {\footnotesize{}Car} & {\footnotesize{}Ped}\tabularnewline
\midrule
{\footnotesize{}Batch Hard Triplet} & {\footnotesize{}76.2} & {\footnotesize{}\textbf{46.8}} & {\footnotesize{}87.8} & {\footnotesize{}\textbf{65.1}} & {\footnotesize{}\textbf{87.2}} & {\footnotesize{}\textbf{75.7}}\tabularnewline
{\footnotesize{}Batch All Triplet} & {\footnotesize{}75.5} & {\footnotesize{}45.3} & {\footnotesize{}87.1} & {\footnotesize{}63.8} & {\footnotesize{}87.1} & {\footnotesize{}75.6}\tabularnewline
{\footnotesize{}Contrastive} & {\footnotesize{}\textbf{76.4}} & {\footnotesize{}43.2} & {\footnotesize{}\textbf{88.7}} & {\footnotesize{}61.5} & {\footnotesize{}86.7} & {\footnotesize{}75.2}\tabularnewline
\bottomrule
\end{tabular}
\par\end{centering}
\caption{\label{tab:results-embedding}\textbf{Different Association Losses for TrackR-CNN}. Comparison of results on the KITTI MOTS validation set.}
\end{table}

\section{Details of the Annotation Procedure}
We noticed that wrong segmentation results often stem from imprecise or wrong bounding box annotations of the original MOT datasets. For example, the annotated bounding boxes for the KITTI tracking dataset \cite{Geiger12CVPR} are amodal, \ie, they extend to the ground even if only the upper body of a person is visible.
In these cases, our annotators corrected these bounding boxes instead of adding additional polygon annotations.
We also corrected the bounding box level tracking annotations in cases where they contained errors or missed objects.
Finally, we retained ignore regions that were labeled in the source datasets, \ie, image regions that contain unlabeled objects from nearby classes (like vans and buses) or target objects that were to small to be labeled.
Hypothesized masks that are mapped to ignore regions are neither counted as true nor as false positives in our evaluation procedure.

\section{Ground Truth Experiments}

\begin{table}[t]
\setlength{\tabcolsep}{4pt}
\begin{centering}
\begin{tabular}{ccccccc}
\toprule
\multirow{2}{*}{} & \multicolumn{2}{c}{{\scriptsize{}sMOTSA}} & \multicolumn{2}{c}{{\scriptsize{}MOTSA}} & \multicolumn{2}{c}{{\scriptsize{}MOTSP}}\tabularnewline
 & {\scriptsize{}Car} & {\scriptsize{}Ped} & {\scriptsize{}Car} & {\scriptsize{}Ped} & {\scriptsize{}Car} & {\scriptsize{}Ped}\tabularnewline
\midrule
{\footnotesize{}GT Boxes (orig) + Filling} & {\footnotesize{}33.7} & {\footnotesize{}-66.1} & {\footnotesize{55.5}} & {\footnotesize{}-57.7} & {\footnotesize{}71.8} & {\footnotesize{}54.6}\tabularnewline
{\footnotesize{}GT Boxes (orig) + Ellipse} & {\footnotesize{}52.3} & {\footnotesize{}-31.9} & {\footnotesize{}74.0} & {\footnotesize{}-14.5} & {\footnotesize{}74.9} & {\footnotesize{}57.4}\tabularnewline
{\footnotesize{}GT Boxes (orig) + MG} & {\footnotesize{}77.3} & {\footnotesize{36.5}} & {\footnotesize{}90.4} & {\footnotesize{}55.7} & {\footnotesize{}86.3} & {\footnotesize{}75.3}\tabularnewline
{\footnotesize{}GT Boxes (tight) + Filling} & {\footnotesize{}61.3} & {\footnotesize{}-1.7} & {\footnotesize{}83.9} & {\footnotesize{}22.0} & {\footnotesize{}75.4} & {\footnotesize{}60.5}\tabularnewline
{\footnotesize{}GT Boxes (tight) + Ellipse} & {\footnotesize{}70.9} & {\footnotesize{}17.2} & {\footnotesize{}91.8} & {\footnotesize{}42.4} & {\footnotesize{}78.1} & {\footnotesize{}64.2}\tabularnewline
{\footnotesize{}GT Boxes (tight) + MG} & {\footnotesize{} \textbf{82.5}} & {\footnotesize{} \textbf{50.0}} & {\footnotesize{} \textbf{95.3}} & {\footnotesize{} \textbf{71.1}} & {\footnotesize{} \textbf{86.9}} & {\footnotesize{} \textbf{75.4}}\tabularnewline
\bottomrule
\end{tabular}
\par\end{centering}
\caption{\label{tab:results-kitti-gt}\textbf{Ground Truth Results on KITTI MOTS}. +MG denotes mask generation with a KITTI MOTS fine-tuned Mask R-CNN..}
\end{table}

We performed additional experiments to demonstrate the difficulty of generating accurate segmentation masks even when the ground truth bounding boxes are given (see Table~\ref{tab:results-kitti-gt}).
As in the main paper, we consider two variants of the ground truth: the original bounding boxes from KITTI (\textit{orig}), which are amodal, \ie if only the upper body of a person is visible, the box will still extend to the ground, and tight bounding boxes (\textit{tight}) derived from our segmentation masks.
We created masks for the boxes by simply filling the full box (\textit{+Filling}), by inserting an ellipse (\textit{+Ellipse}), and by generating masks using the KITTI MOTS fine-tuned Mask R-CNN (\textit{+MG}). In each case, instance ids are retained from the corresponding boxes.

Our results show that rectangles and ellipses are not sufficient to accurately localize objects when mask-based matching is used%
, even with perfect track hypotheses. The problem is amplified when using amodal boxes, which often contain large regions that do not show the object. This further validates our claim that MOT tasks can benefit from pixel-wise evaluation. The relatively low scores for pedestrians also imply a limit to post-hoc masks generation using the KITTI fine-tuned Mask R-CNN.

\section{Visualization of Association Vectors}
\begin{figure*}[t!]
	\centering
	\includegraphics[width=\textwidth]{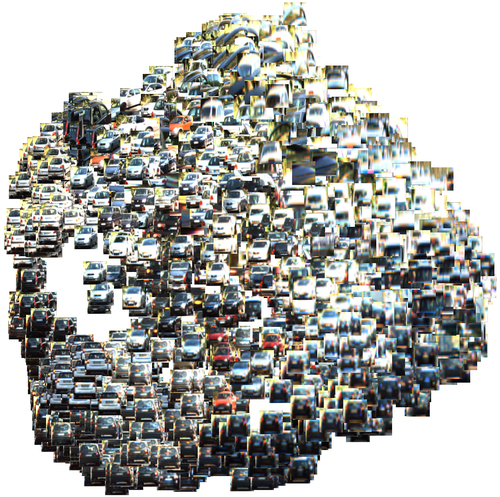}
	\caption{\textbf{Visualization using PCA on the association vectors of detections generated by TrackR-CNN on sequence 18 of KITTI MOTS.} Detections with similar appearance are grouped together by minimizing the association loss.}
	\label{fig:trackr-cnn-embedding-pca}
\end{figure*}

We present a visualization of the association vectors produced by our TrackR-CNN model on a sequence of the KITTI MOTS validation set in Figure~\ref{fig:trackr-cnn-embedding-pca}. Here, all association vectors for detections produced by TrackR-CNN on sequence 18 were used for principal component analysis and then projected onto the two components explaining most of their variance. The resulting two dimensional vectors were used to arrange the crops for the corresponding detections in 2D. The visualization was created using the TensorBoard embedding projector. It can be seen that crops belonging to the same car are in most cases close to each other in the embedding space.

\section{Qualitative results}
We present further qualitative results of our baseline TrackR-CNN model on the KITTI MOTS and MOTSChallenge validation sets including some illustrative failure cases. See Figures~\ref{fig:mots-results1}, \ref{fig:kitti-results1}, \ref{fig:kitti-results3} and \ref{fig:kitti-results4} on the following pages.

\newcommand{\imgmots}[1] {\includegraphics[width=0.23\textwidth]{figures/qualitative/MOTSChallenge/0005/#1}}  
\begin{figure*}[t!]
	\centering
		\imgmots{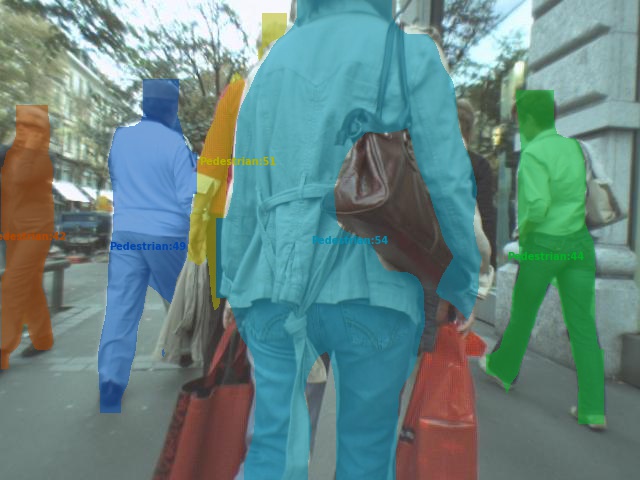}
		\vspace{1.2pt}
		\imgmots{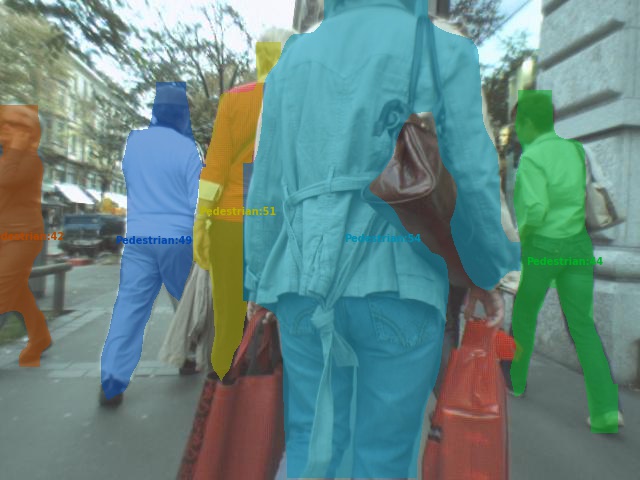}
		\imgmots{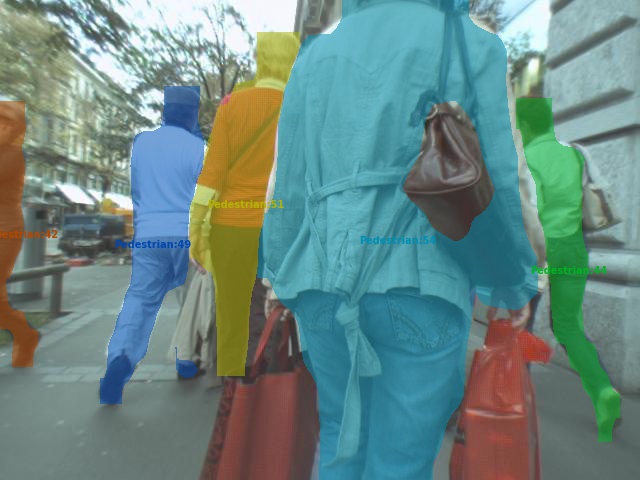}
		\imgmots{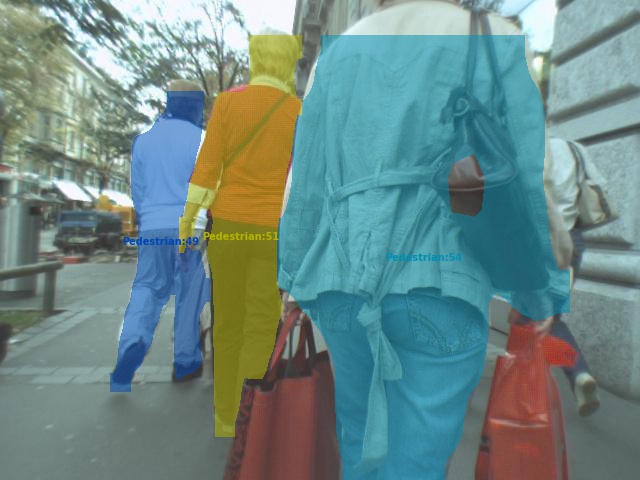}
		\\
		
		\imgmots{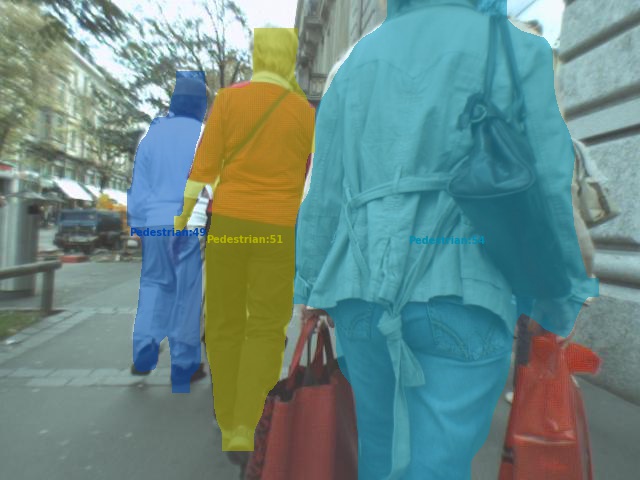}
		\vspace{1.2pt}
		\imgmots{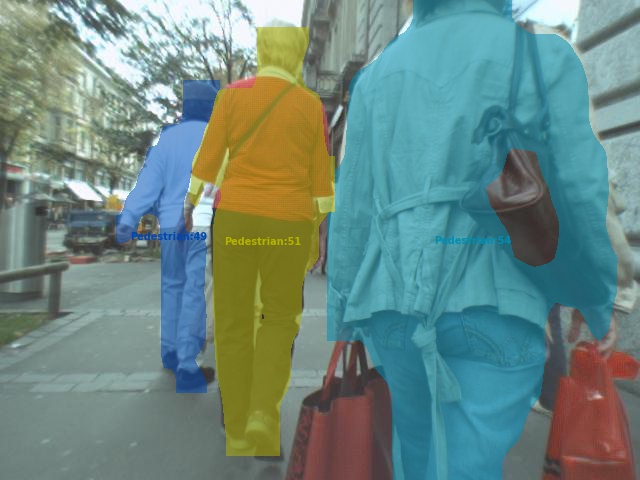}
		\imgmots{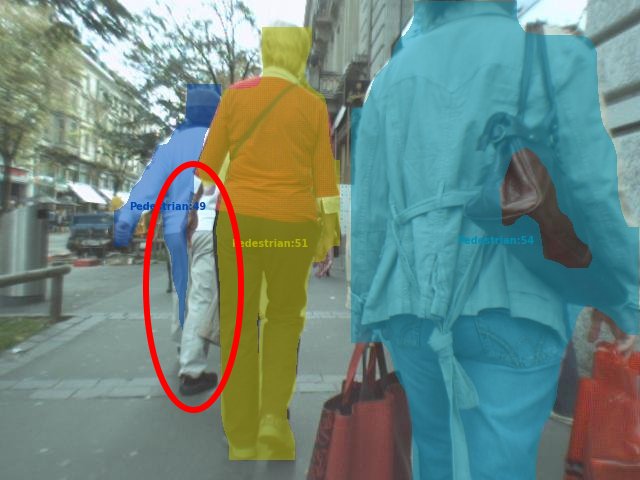}
		\imgmots{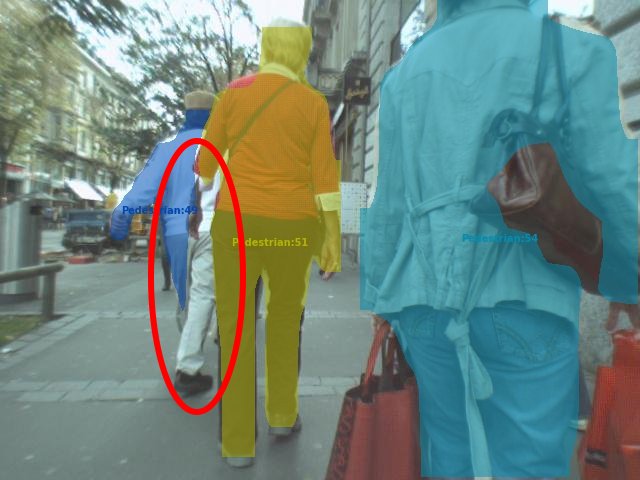}
		\\
		
		\imgmots{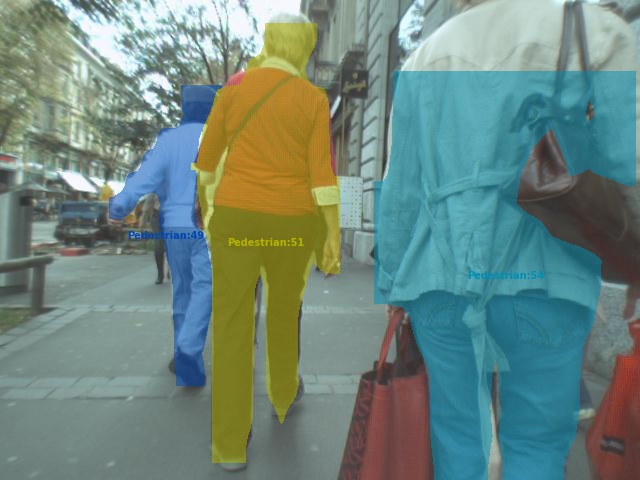}
		\vspace{1.2pt}
		\imgmots{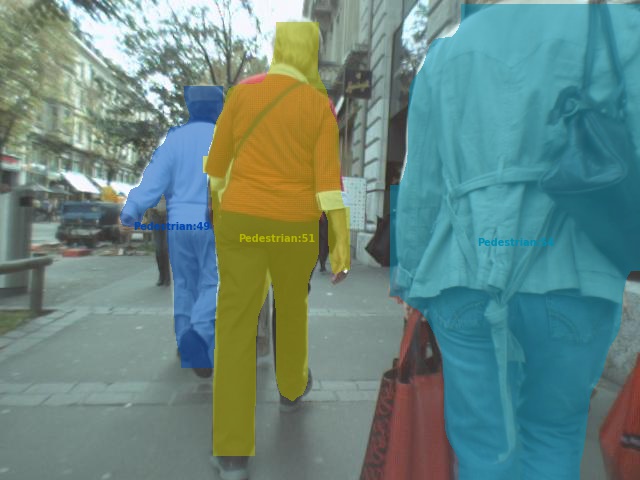}
		\imgmots{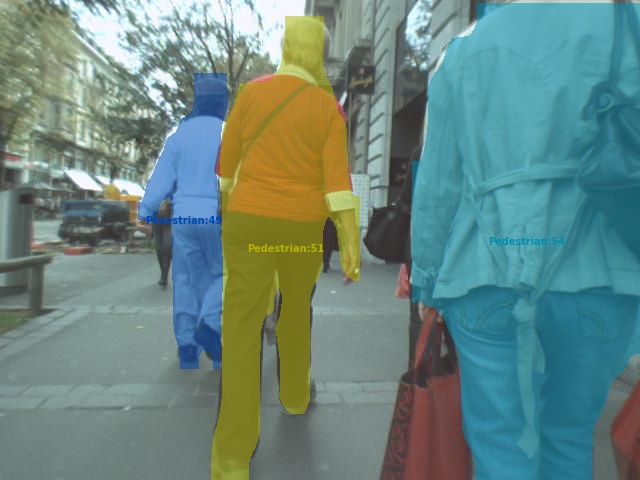}
		\imgmots{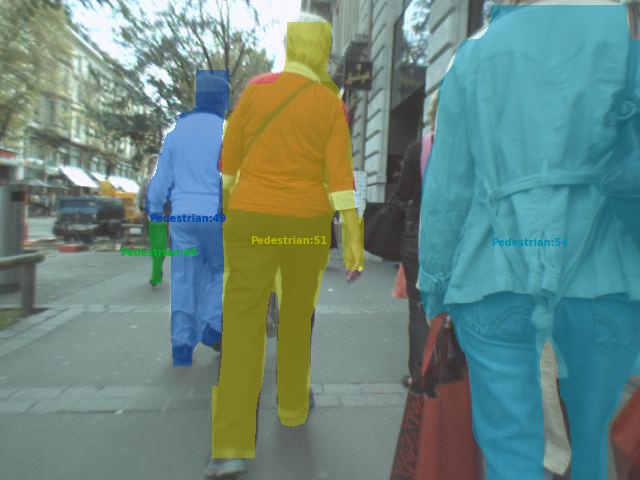}
		\\
		
		\imgmots{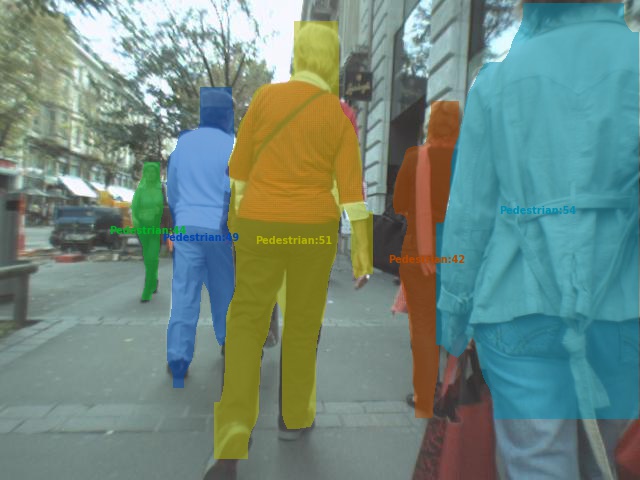}
		\vspace{1.2pt}
		\imgmots{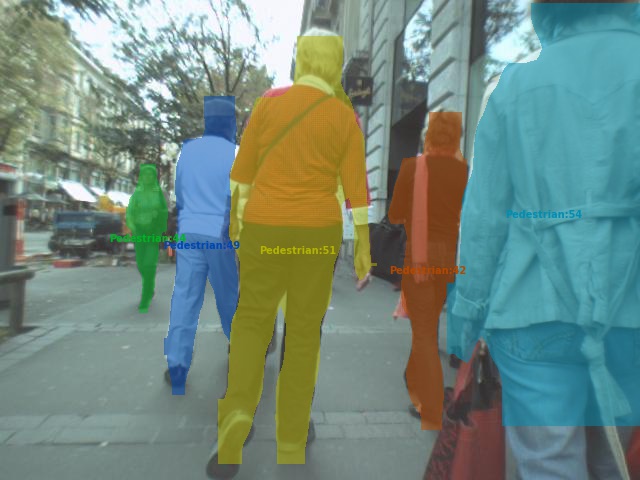}
		\imgmots{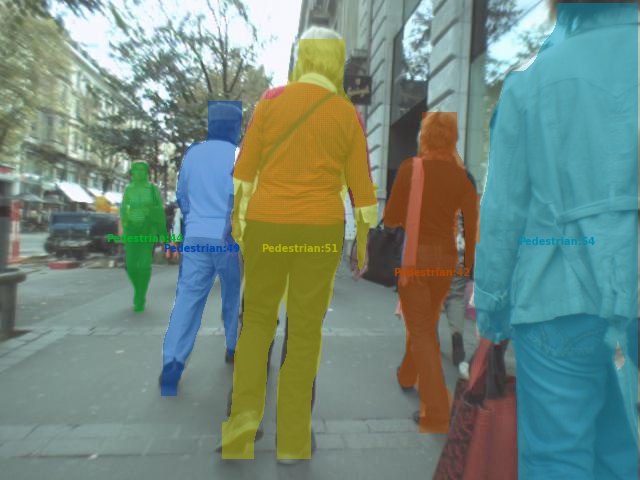}
		\imgmots{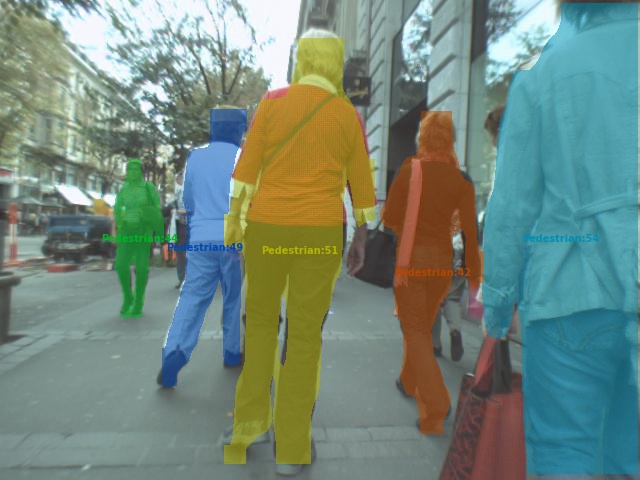}
		\\

		\includegraphics[width=0.23\textwidth]{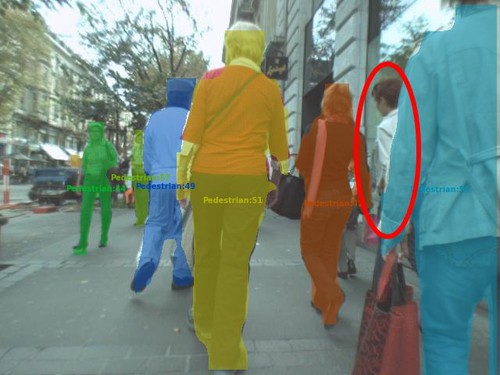}
		\vspace{1.2pt}
		\imgmots{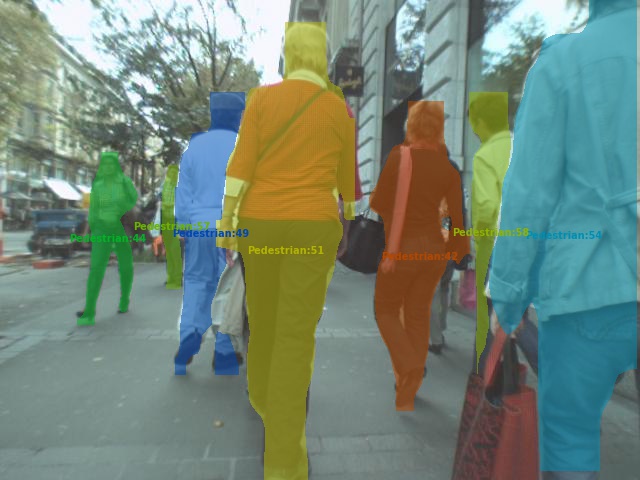}
		\imgmots{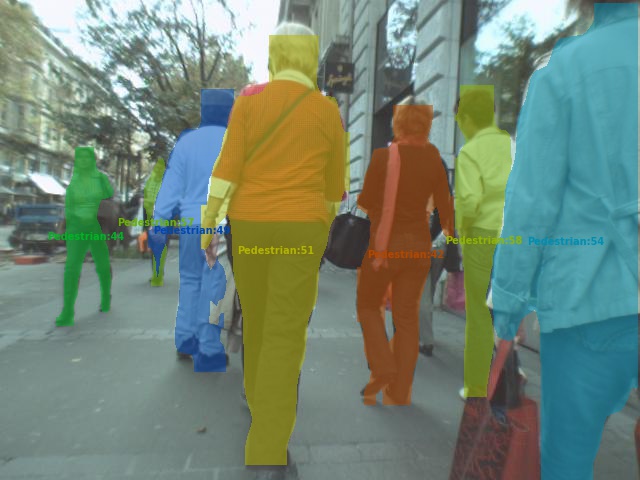}
		\imgmots{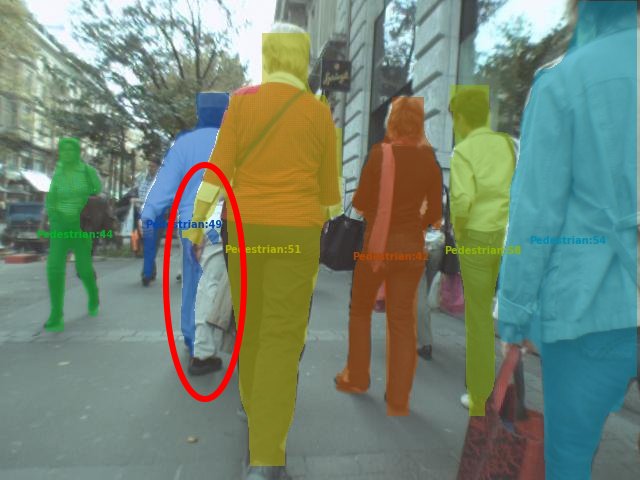}
		\\

		\includegraphics[width=0.23\textwidth]{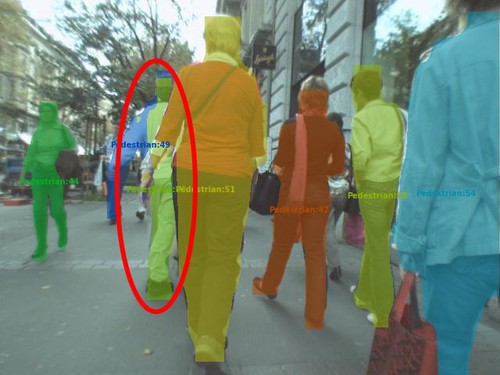}
		\vspace{1.2pt}
		\includegraphics[width=0.23\textwidth]{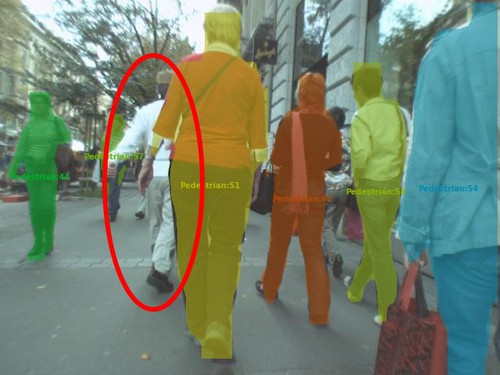}
		\includegraphics[width=0.23\textwidth]{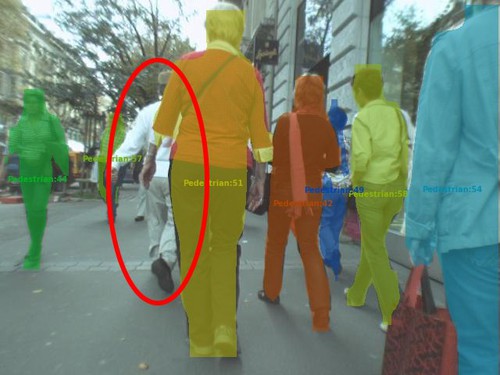}
		\imgmots{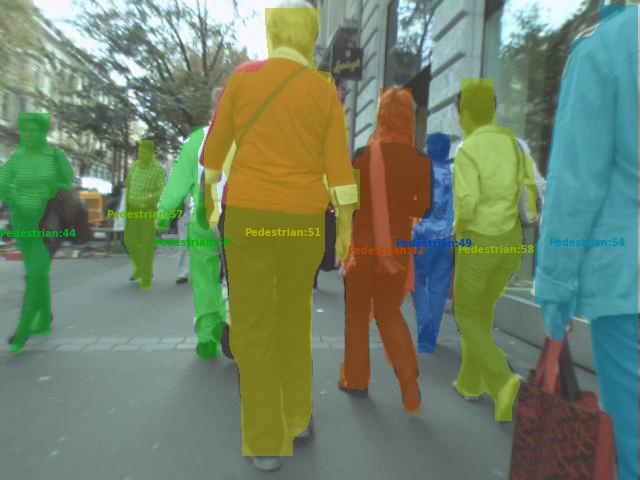}
		\\
    \vspace{-6pt}%
	\caption{\textbf{Qualitative Results on MOTSChallenge.} While complex scenes with many occluding objects often work well, there can still be missing detections and id switches during difficult occlusions, as in this example (highlighted by red ellipses).
	}
	\label{fig:mots-results1}
\end{figure*}

\newcommand{\imgkittia}[1] {\includegraphics[width=0.49\textwidth]{figures/qualitative/KITTI_MOTS/0006/#1}}  
\begin{figure*}[t!]
	\centering
		\imgkittia{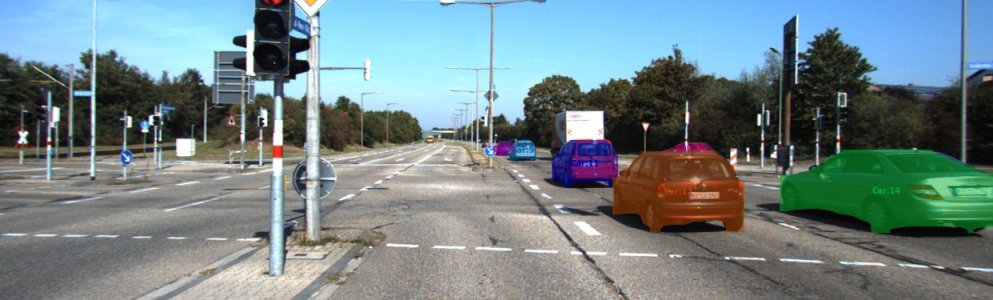}
		\vspace{1.2pt}
		\imgkittia{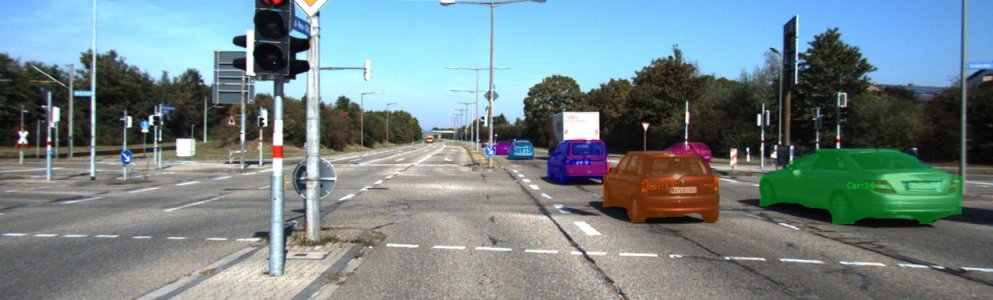}
		\\

		\imgkittia{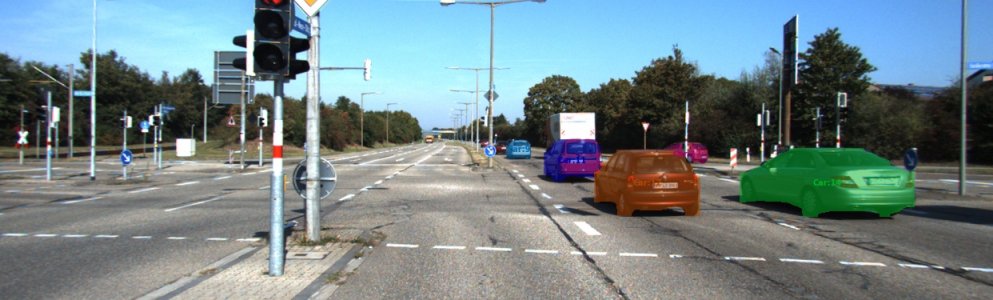}		
		\vspace{1.2pt}
		\imgkittia{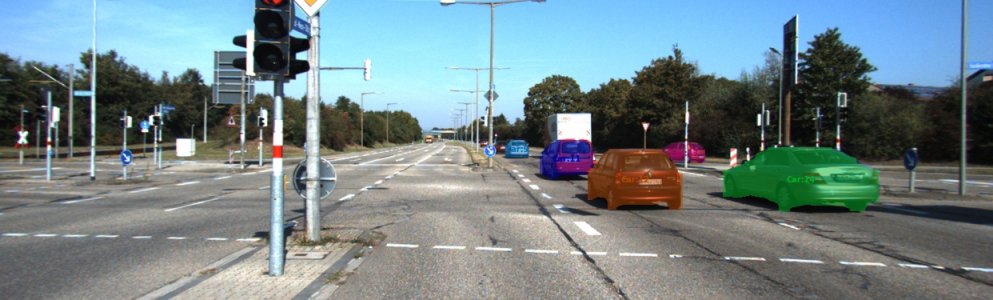}
		\\
		
		\imgkittia{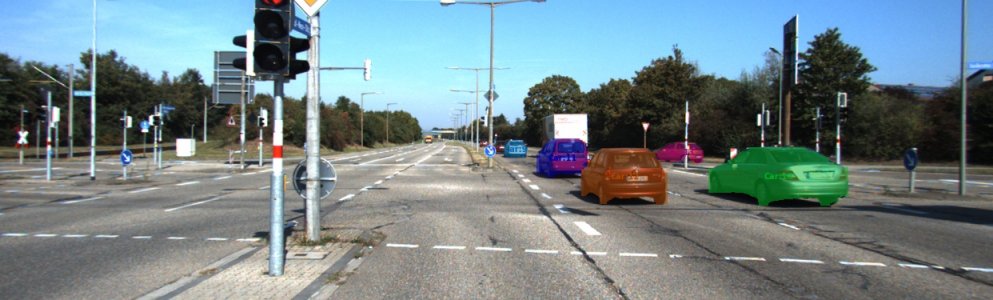}
 		\vspace{1.2pt}
		\imgkittia{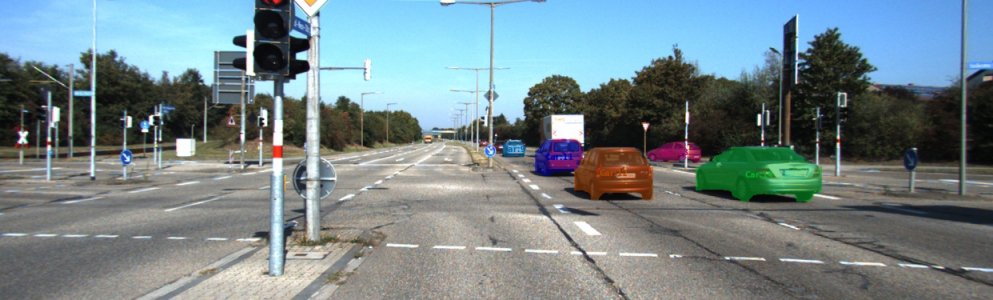}
		\\
		
		\imgkittia{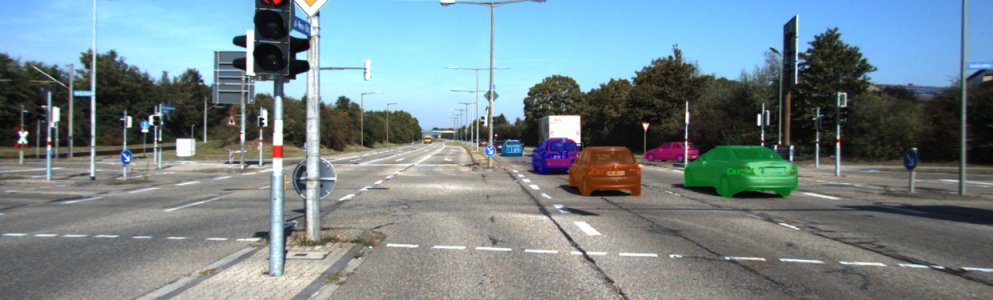}
		\vspace{1.2pt}
		\includegraphics[width=0.49\textwidth]{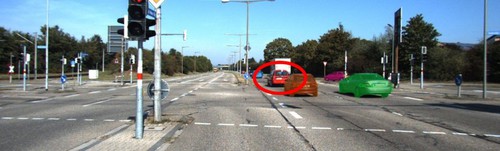}
		\\

		\imgkittia{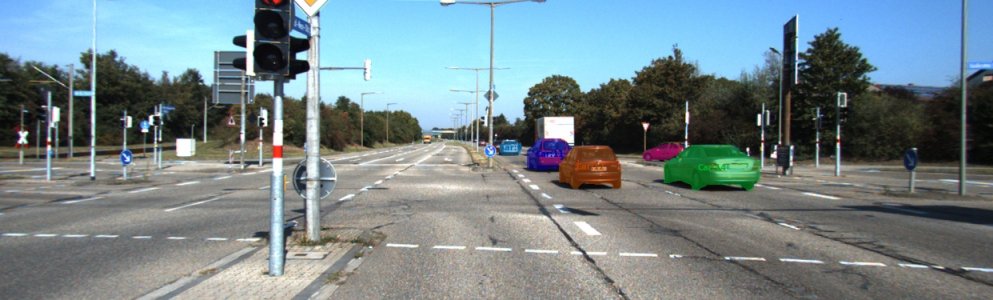}		
		\vspace{1.2pt}
		\imgkittia{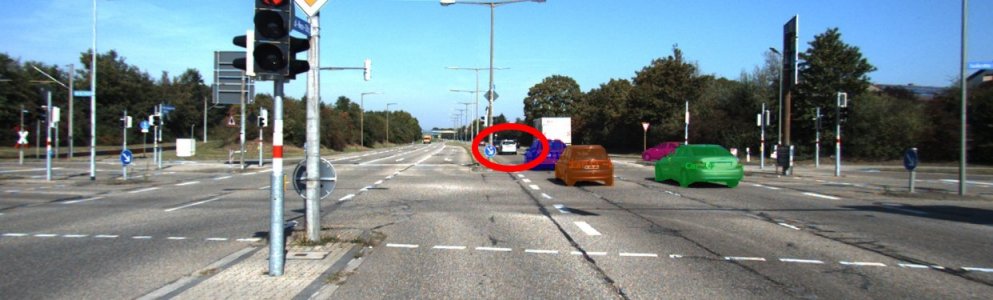}
		\\

		\imgkittia{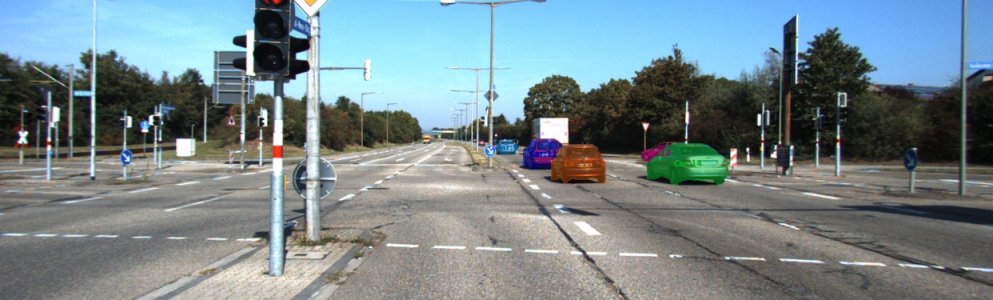}
		\vspace{1.2pt}
		\imgkittia{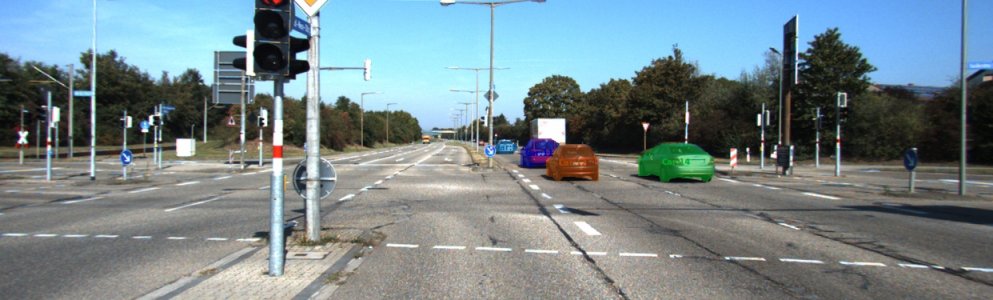}
		\\
		
		\imgkittia{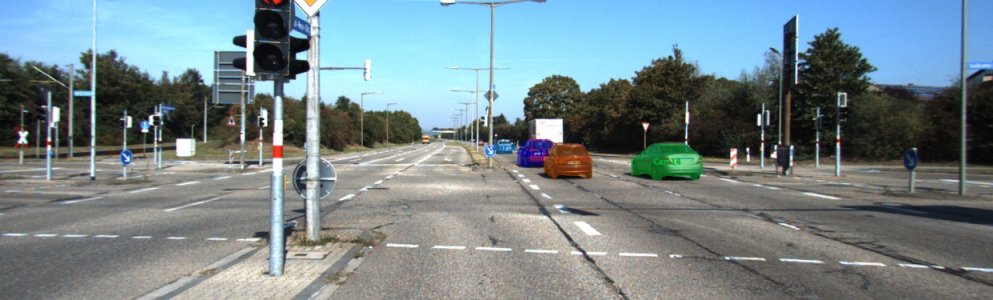}
		\vspace{1.2pt}
		\imgkittia{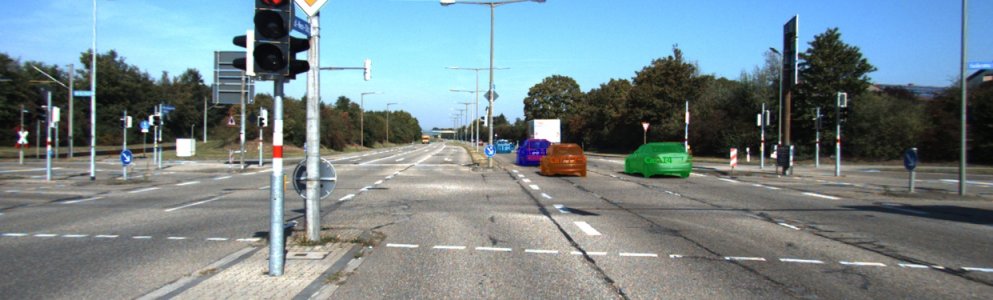}
    \vspace{-6pt}%
	\caption{\textbf{Qualitative Results on KITTI MOTS.} In simpler scenes, the model is able to continue a track with the same ID after a missing detection (highlighted by red ellipses).
	}
	\label{fig:kitti-results1}
\end{figure*}

\newcommand{\imgkittic}[1] {\includegraphics[width=0.49\textwidth]{figures/qualitative/KITTI_MOTS/0008/#1}}  
\begin{figure*}[t!]
	\centering
		\includegraphics[width=0.49\textwidth]{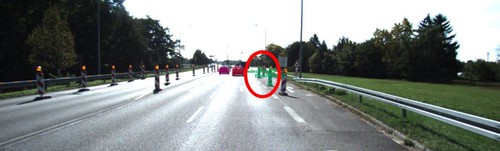}
		\vspace{1.2pt}
		\imgkittic{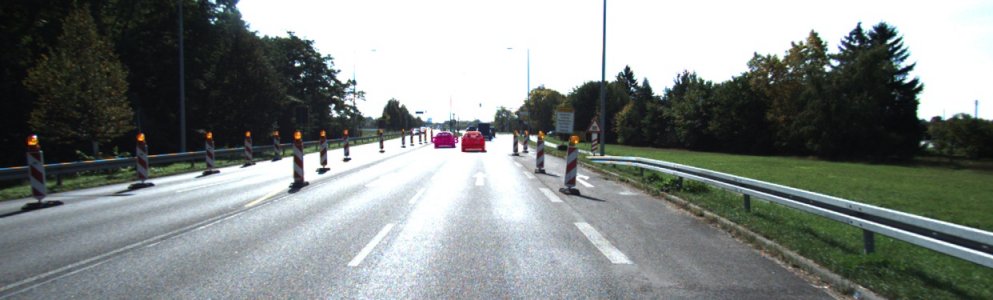}
		\\
		
		\imgkittic{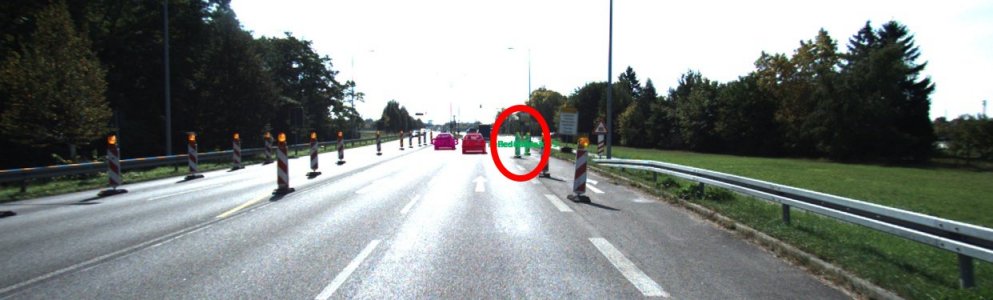}		
		\vspace{1.2pt}
		\includegraphics[width=0.49\textwidth]{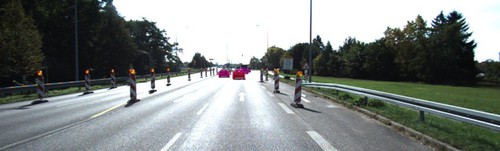}
		\\
		
		\imgkittic{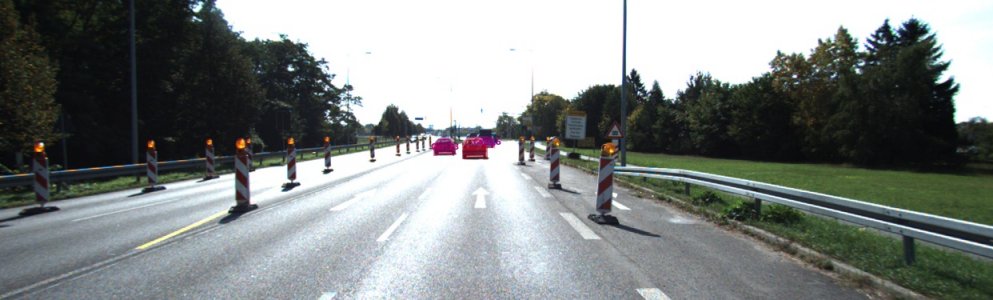}
		\vspace{1.2pt}
		\includegraphics[width=0.49\textwidth]{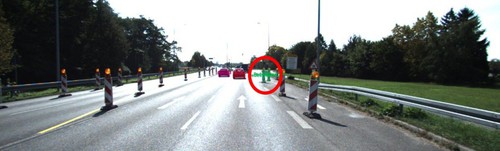}
	    \\
	    
	    \includegraphics[width=0.49\textwidth]{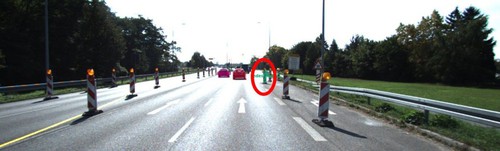}
   		\vspace{1.2pt}
   		\includegraphics[width=0.49\textwidth]{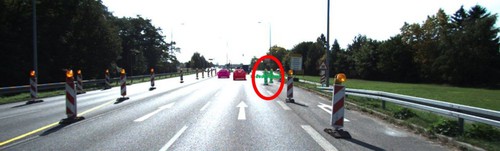}
    \vspace{-6pt}%
	\caption{\textbf{Qualitative Results on KITTI MOTS.} In a rare failure case, pylons are confused for pedestrians (highlighted by red ellipses). In most cases, detections correspond to real instances of the class.
	}
	\label{fig:kitti-results3}
\end{figure*}

\newcommand{\imgkittid}[1] {\includegraphics[width=0.49\textwidth]{figures/qualitative/KITTI_MOTS/0013/#1}}  
\begin{figure*}[t!]
	\centering
		\imgkittid{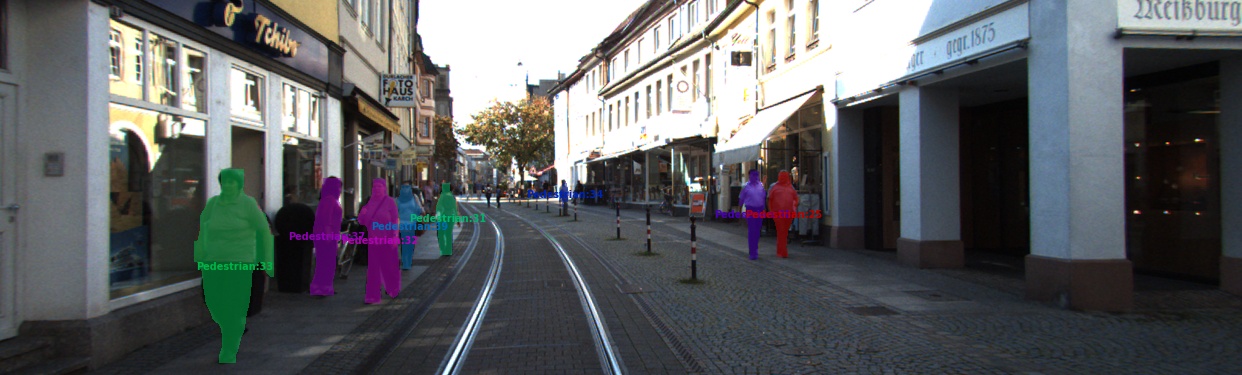}
		\vspace{1.2pt}
		\imgkittid{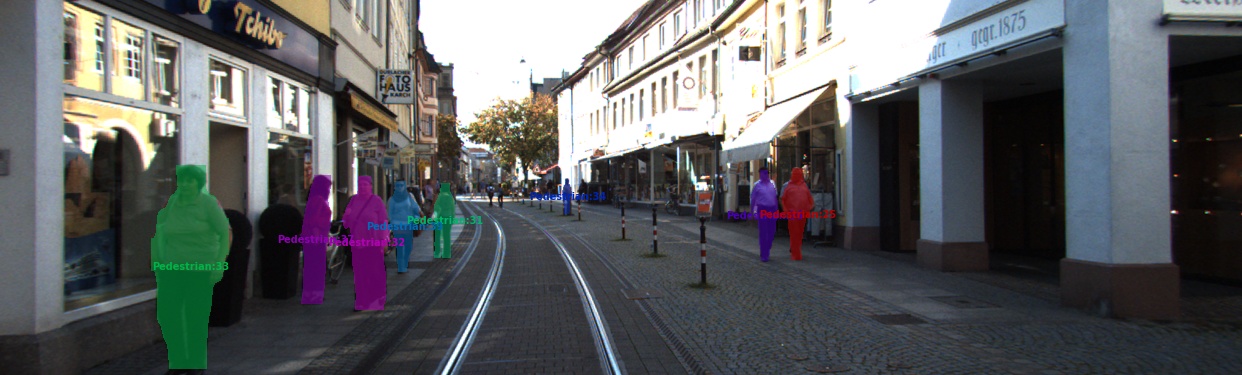}
		\\
		
		\imgkittid{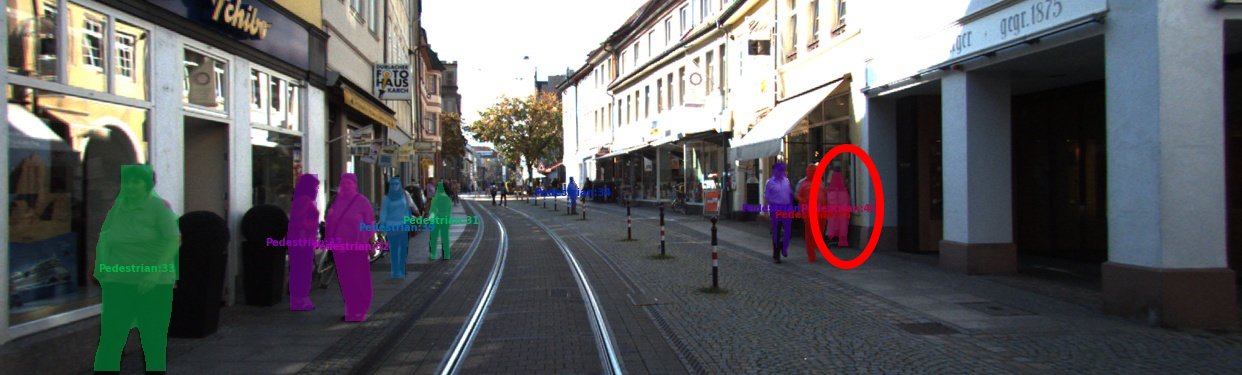}
		\vspace{1.2pt}		
		\imgkittid{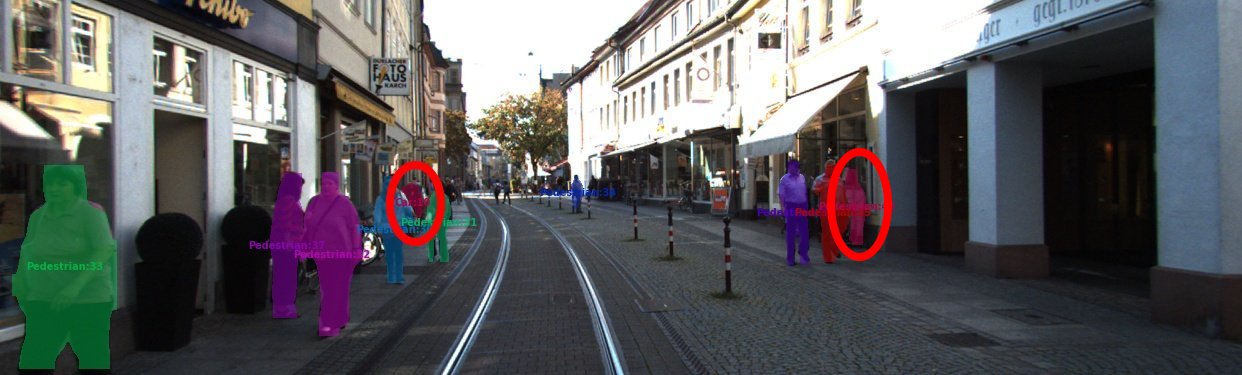}
		\\
		
		\imgkittid{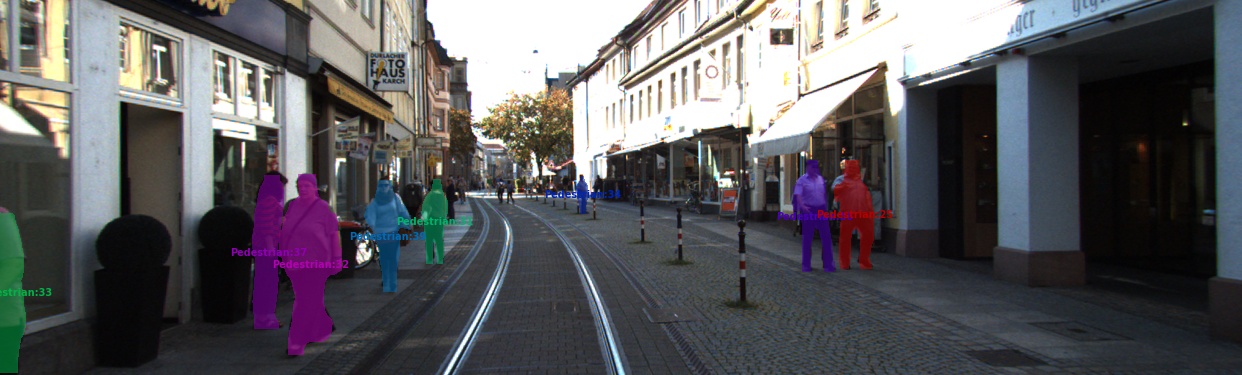}
		\vspace{1.2pt}
		\imgkittid{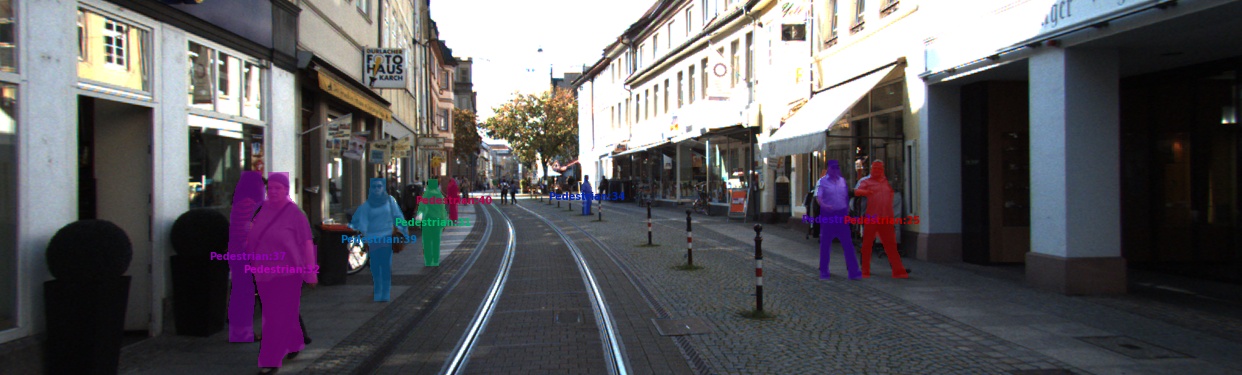}
		\\
		
		\imgkittid{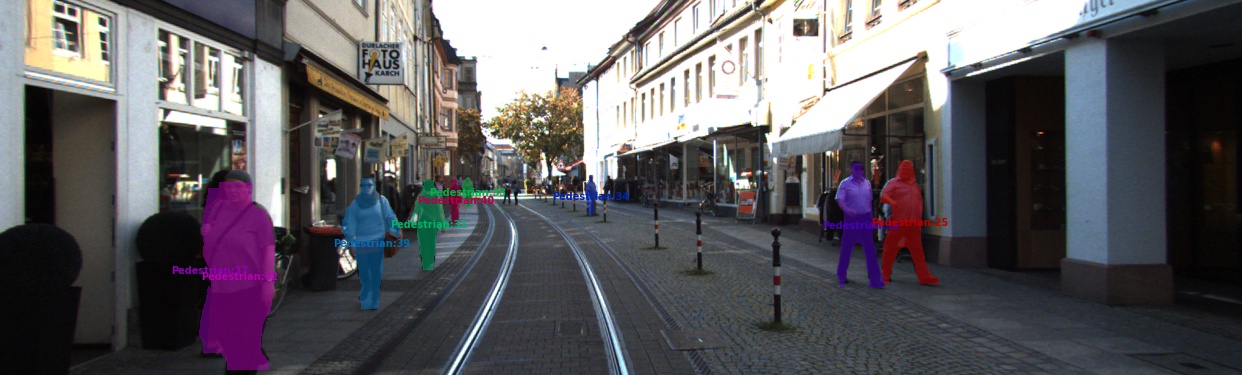}
		\vspace{1.2pt}
		\imgkittid{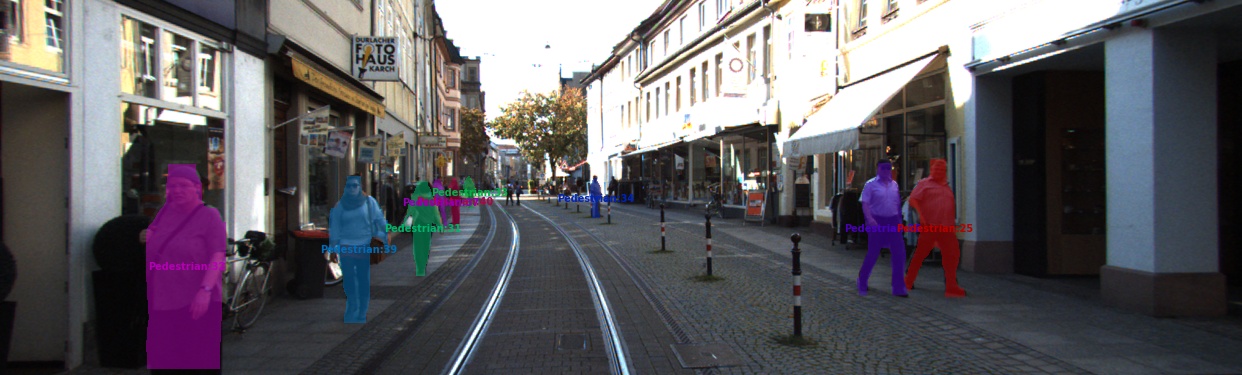}
		\\
		
		\imgkittid{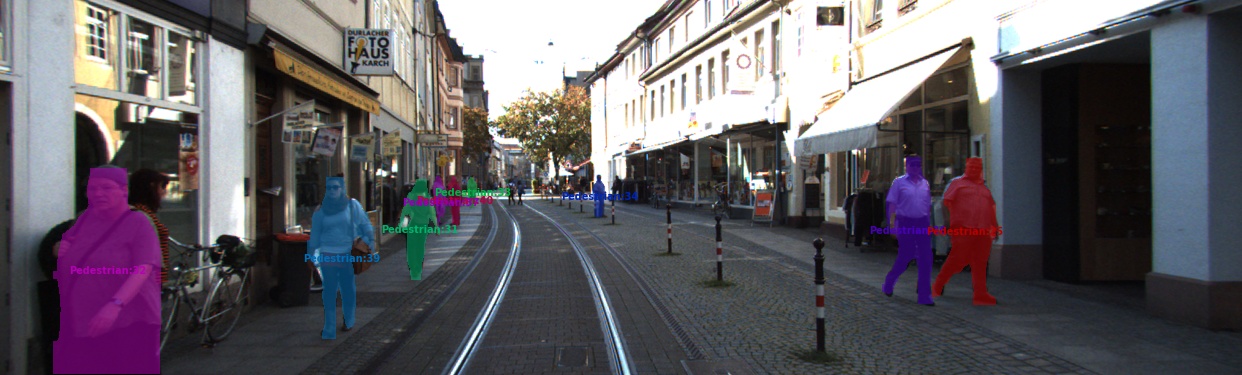}
		\vspace{1.2pt}		
		\imgkittid{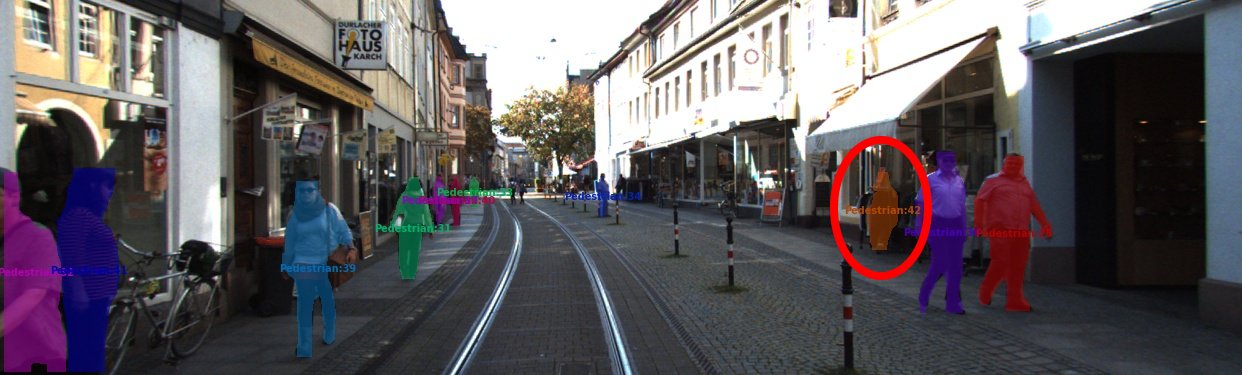}
		\\

		\imgkittid{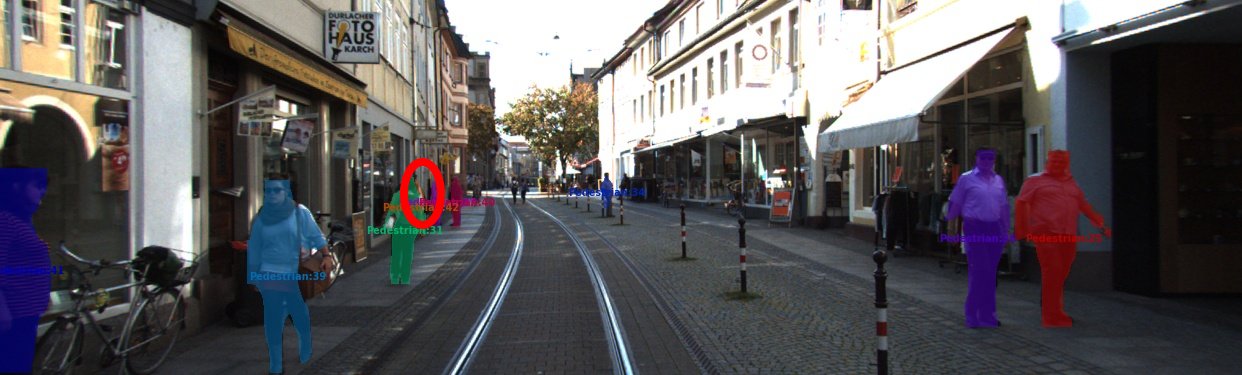}
		\vspace{1.2pt}
		\imgkittid{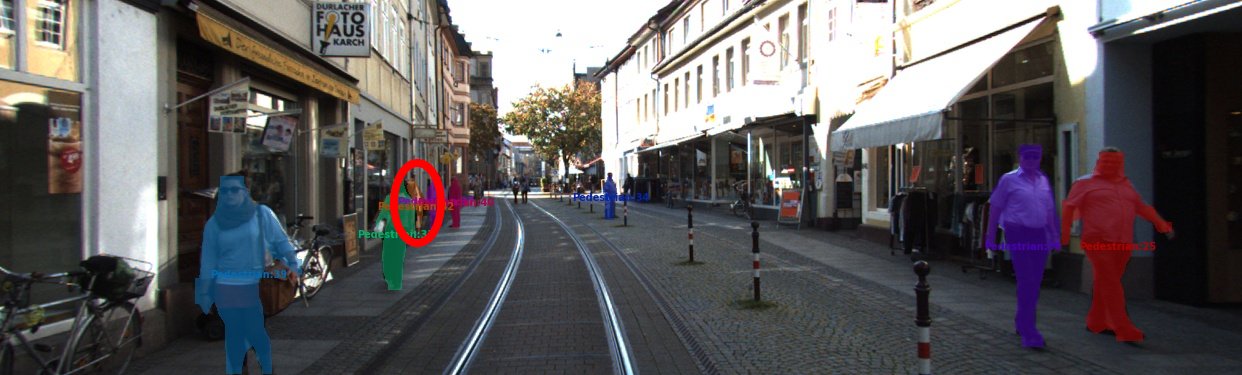}
		\\
		
		\imgkittid{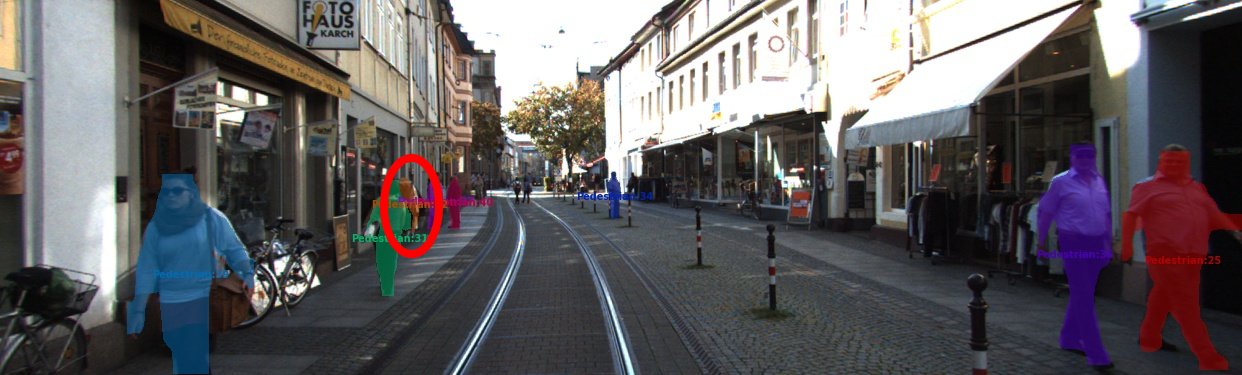}
		\vspace{1.2pt}
		\includegraphics[width=0.49\textwidth]{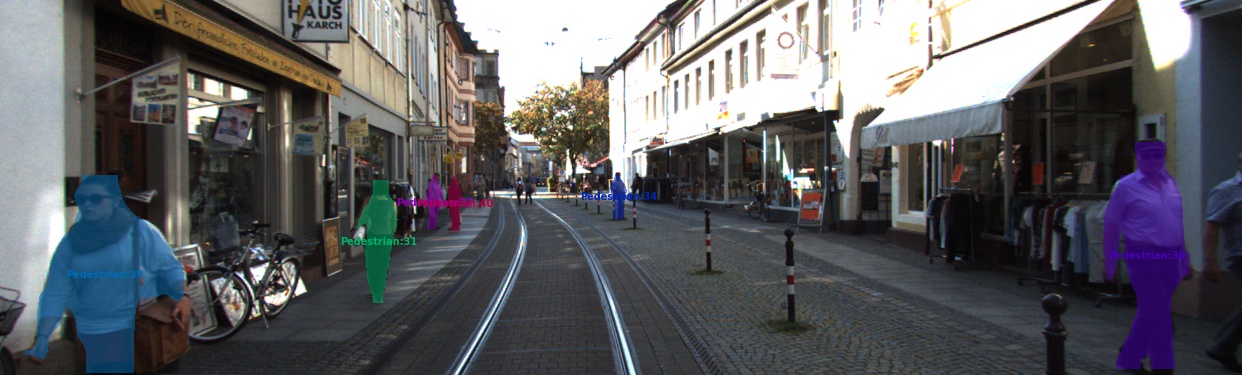}
		\\
		
		\imgkittid{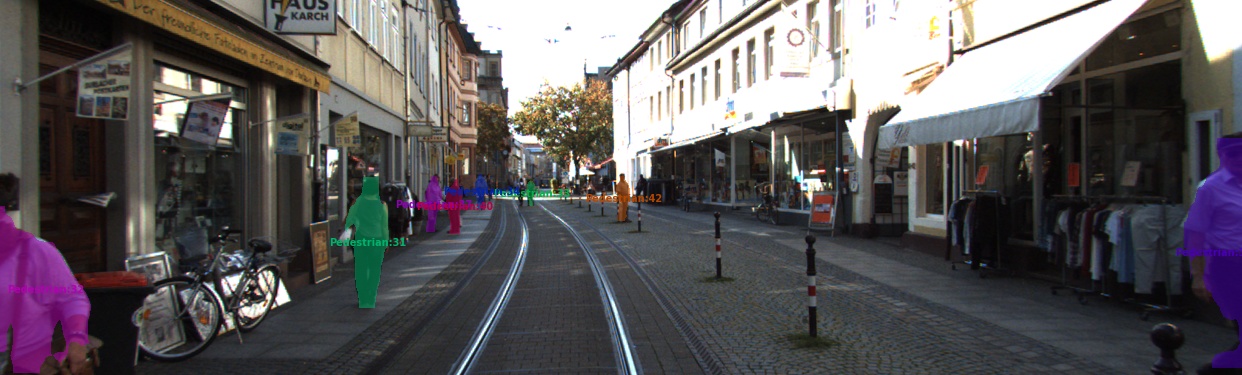}
		\vspace{1.2pt}
		\imgkittid{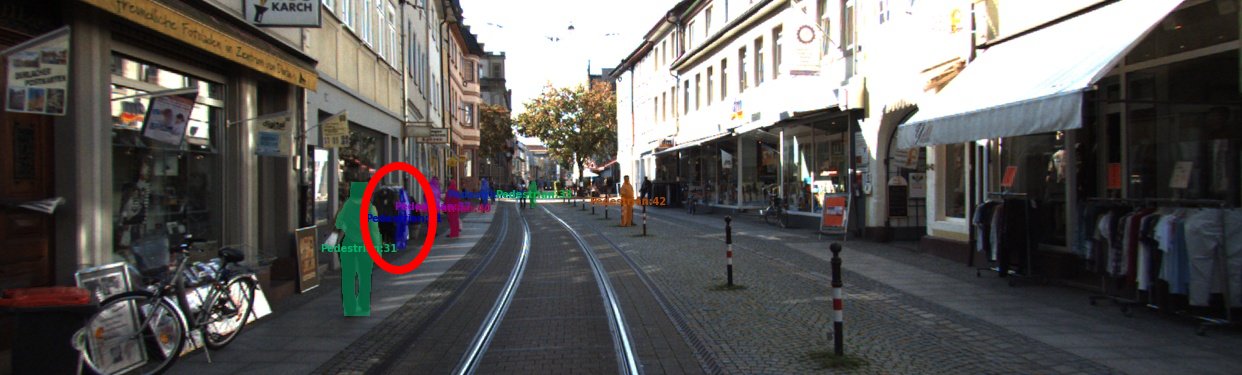}
    \vspace{-6pt}%
	\caption{\textbf{Qualitative Results on KITTI MOTS.} In less crowded scenes, distinguishing objects works well but some erroneous detections (highlighted by red ellipses) might still happen.
	}
	\label{fig:kitti-results4}
\end{figure*}

\end{document}